\documentclass[final,5p,times,twocolumn]{elsarticle}

\usepackage{amsmath}
\usepackage{mathtools}
\usepackage{xcolor}
\usepackage{enumitem}
\usepackage{comment}
\usepackage{graphicx}
\usepackage{multirow}
\usepackage{array}
\usepackage{caption}
\usepackage{fontawesome5}
\usepackage{hyperref}
\usepackage{wrapfig}
\usepackage{graphicx}
\usepackage{caption}
\usepackage{booktabs}
\usepackage{adjustbox}
\usepackage{footnote}
\makesavenoteenv{table}
\usepackage{caption}
\usepackage{multicol}

\usepackage{graphicx}
\usepackage[font=small,labelfont=bf]{caption}

\usepackage{subcaption}

\usepackage{amssymb}

\usepackage{booktabs}
\usepackage{titlesec}

\titlespacing\section{0pt}{10pt plus 4pt minus 2pt}{0pt plus 2pt minus 2pt}
\titlespacing\subsection{0pt}{8pt plus 4pt minus 2pt}{0pt plus 2pt minus 2pt}

\begin{document}

\begin{frontmatter}



\title{ 
Explainable AI: Context-Aware Layer-Wise Integrated Gradients for Explaining Transformer Models}

\author[1]{Melkamu Abay Mersha} 
   
\author[1]{Jugal Kalita}

\affiliation[1]{organization={College of Engineering and Applied Science, University of Colorado Colorado Springs},
            addressline={Colorado Springs},
            postcode={80918},
            state={CO},
            country={USA}}

\cortext[cor1]{Corresponding author.}  

\cortext[email]{\textit{E-mail addresses:} 
\href{mailto:melkamu.mersha@uccs.edu}{mmersha@uccs.edu} (M.A. Mersha*), 
\href{mailto:jkalita@uccs.edu}{jkalita@uccs.edu} (J. Kalita).}

\begin{abstract}
Transformer models achieve state-of-the-art performance across domains and tasks, yet their deeply layered representations make their predictions difficult to interpret. Existing explainability methods rely on final-layer attributions, capture either local token-level attributions or global attention patterns without unification, and lack context-awareness of inter-token dependencies and structural components. They also fail to capture how relevance evolves across layers and how structural components shape decision-making. To address these limitations, we proposed the \textbf{Context-Aware Layer-wise Integrated Gradients (CA-LIG) Framework}, a unified hierarchical attribution framework that computes layer-wise Integrated Gradients within each Transformer block and fuses these token-level attributions with class-specific attention gradients. This integration yields signed, context-sensitive attribution maps that capture supportive and opposing evidence while tracing the hierarchical flow of relevance through the Transformer layers.
We evaluate the CA-LIG Framework across diverse tasks, domains, and transformer model families, including sentiment analysis and long and multi-class document classification with BERT, hate speech detection in a low-resource language setting with XLM-R and AfroLM, and image classification with Masked Autoencoder vision Transformer model. Across all tasks and architectures, CA-LIG provides more faithful attributions, shows stronger sensitivity to contextual dependencies, and produces clearer, more semantically coherent visualizations than established explainability methods. These results indicate that CA-LIG provides a more comprehensive, context-aware, and reliable explanation of Transformer decision-making, advancing both the practical interpretability and conceptual understanding of deep neural models.\\
{\footnotesize The implementation code will be made publicly available at \url{https://github.com/melkamumersha/Context-Aware-XAI} upon acceptance of the paper.}

\end{abstract}

\begin{keyword}
explainable artificial intelligence (XAI), interpretable deep learning,  large language models (LLMs), natural language processing (NLP), explainability techniques, Vision Transformers, computer vision, black-box models.
\end{keyword}

\end{frontmatter}

\makeatletter
\def\ps@pprintTitle{%
    \let\@oddhead\@empty
    \let\@evenhead\@empty
    \def\@oddfoot{\hbox to \textwidth{\hfil\thepage\hfil}}%
    \let\@evenfoot\@oddfoot
}
\makeatother

\section{Introduction}
Transformer-based architectures, such as BERT~\cite{devlin2018bert}, 
GPT~\cite{radford2018improving}, and T5~\cite{raffel2020exploring}, have become 
foundational across modern natural language processing (NLP) and a wide range 
of artificial intelligence applications \cite{mersha2024semantic, khapre2025toxicity, tonja2023first}. Built upon multi-head self-attention~\cite{vaswani2017attention}, 
deep hierarchical representations, and scalable parallel computation, Transformers 
model long-range dependencies and complex semantic interactions with exceptional 
effectiveness. Their architectural components---including self-attention, feedforward 
sublayers, residual connections, positional encodings, and normalization layers---jointly 
yield rich contextual representations that support tasks such as sentiment analysis, 
text classification, multimodal reasoning, and image recognition~\cite{rogers2020primer, mersha2025semantic}.

Despite their success, Transformer models remain fundamentally opaque. Their layered 
and nonlinear structure makes it difficult to determine how token representations evolve 
across layers, how contextual dependencies influence predictions, and how evidence flows 
through the network. This inherent opacity has intensified the need for eXplainable AI 
(XAI) techniques capable of producing faithful and interpretable insights into Transformer 
decision-making.

A wide spectrum of explanation approaches has been proposed, yet each exhibits notable 
limitations. \textit{Attention-based methods} treat attention distributions as indicators of token importance 
~\cite{liu2021exploring, yeh2023attentionviz}, but numerous studies show that raw attention weights do not reliably reflect model reasoning or provide faithful explanations~\cite{jain2019attention, serrano2019attention}. Extensions such as 
Attention Rollout or attention-flow~\cite{abnar2020quantifying} analyses improve 
traceability across layers but still rely solely on attention weights, omitting contributions 
from feedforward networks, value vectors, residual pathways, and normalization layers \cite{alshami2025smart}.

\textit{Gradient-based approaches}, including Integrated Gradients (IG)~\cite{sundararajan2017axiomatic} 
and Guided IG~\cite{kapishnikov2021guided}, provide theoretically sound input-level 
attributions but are typically computed only at the final layer, failing to capture how 
relevance evolves across intermediate Transformer blocks. They also struggle to incorporate 
structural relationships between tokens---an essential component of Transformer reasoning.

\textit{Activation-based techniques} such as Layer-wise Relevance Propagation (LRP)~\cite{arras2017explaining} 
propagate relevance through network layers but face difficulty preserving relevance 
conservation across multi-head attention, often producing coarse and context-insensitive 
attributions~\cite{lan2025attention}.

Overall, existing XAI methods suffer from three fundamental limitations. First, most techniques exhibit a \textit{final-layer bias}, generating explanations only at the final prediction layer while overlooking how semantic information and contextual abstractions are progressively formed across earlier layers of the model. Second, current approaches lack \textit{unified local--global reasoning}: they typically capture either local token-level salience, as in gradient-based methods such as Integrated Gradients, or global structural interactions, as seen in attention-based techniques like attention flow, but rarely integrate both perspectives into a single coherent explanatory representation. Third, prevailing methods offer \textit{insufficient context awareness}, as they often fail to account for inter-token dependencies, residual connections, feedforward transformations, and cross-layer information flow, all of which are
central to the Transformer architecture.
These limitations underscore the need for more holistic, context-aware explanation frameworks that reflect the hierarchical and interconnected nature of Transformer language models.

To address these limitations, we proposed the \textbf{Context-Aware Layer-wise Integrated Gradients (CA-LIG) Framework}, a unified, hierarchical framework that produces faithful, context-sensitive explanations for Transformer-based models. 

Prior studies have shown that raw attention weights do not reliably reflect model reasoning or yield faithful explanations. In CA-LIG, attention is not treated as a faithful explanation mechanism nor used in isolation to explain model decisions. Instead, Integrated Gradients serves as the primary attribution method, providing relevance scores that satisfy axiomatic properties such as completeness and sensitivity. Attention gradients are incorporated solely as a contextual interaction signal, capturing how relevance propagates across tokens and layers within the Transformer architecture. Faithfulness is grounded in IG, while attention gradients refine relevance distribution by accounting for inter-token and inter-layer dependencies.

Rather than concentrating attribution solely at the model’s final layer, the CA-LIG Framework computes Layer-wise Integrated Gradients (LIG) at every Transformer block, capturing how token relevance evolves as representations move through the model hierarchy. These layer-wise relevance signals are then fused with class-specific attention gradients, enabling the framework to reflect both local token contributions and global structural dependencies. The resulting attribution maps are signed, relevance-conserving, and sensitive to contextual interactions, capturing supportive and opposing evidence that aligns with the model’s actual internal reasoning process.

Our contributions are summarized as follows:
\begin{itemize}
    \item We propose a unified and hierarchical XAI framework that captures how token relevance evolves across Transformer layers, enabling layer-wise interpretability rather than restricting explanations to the final output layer.

    \item We design an integrated gradient–attention attribution mechanism that fuses layer-wise gradients with attention-gradient structures, bridging local token relevance with global interaction patterns.

    \item We develop a context-aware XAI framework that enforces normalization and relevance preservation across multi-head attention pathways, improving interpretability.

    \item We conduct a comprehensive empirical evaluation that includes qualitative and quantitative assessments.

    \item We demonstrate the cross-domain and task generality of our framework by validating it across NLP tasks using BERT, XLM-R, AfroLM, and MAE Vision Transformers for the vision task.
\end{itemize}

The remainder of this paper is organized as follows. Section~\ref{sec:related} reviews related work on Transformer explainability. Section~\ref{sec:method} introduces the CA-LIG framework, including its layer-wise decomposition and attention–gradient fusion. Section~\ref{sec:experiments} describes the experimental setup, datasets, baselines, and evaluation metrics. Section~\ref{sec:results} presents the empirical results, including qualitative, quantitative, and causal analyses. Limitations and future work are discussed in Section~\ref{sec:limitations}. Finally, Section~\ref{sec:conclusion} concludes the paper.

\section{Related Work} \label{sec:related}

\subsection{Explainability Techniques}
 A wide range of explainability methods have been developed to interpret NLP models by identifying the importance of individual input tokens to a model’s decision \cite{mersha2024ethio}. These methods are generally grouped into five main categories: perturbation-based, activation-based, gradient-based, attention-based, and hybrid approaches~\cite{mersha2025explainable, mersha2025unified}.

\textit{Perturbation-based} methods work by systematically modifying input features and observing the resulting change in the model’s output \cite{mersha2024explainability}. This class of techniques is model-agnostic and offers intuitive insights. For example, SHAP (SHapley Additive exPlanations) applies game-theoretic principles to assign importance scores to input features~\cite{lundberg2017unified}. LIME (Local Interpretable Model-Agnostic Explanations) builds a local surrogate model around a specific prediction to explain its rationale~\cite{ribeiro2016should, kamen2025introducing}. Occlusion Sensitivity identifies important regions by masking parts of the input and analyzing the impact on predictions~\cite{zeiler2014visualizing}.

\textit{Activation-based} approaches rely on analyzing internal neuron activations to trace how input features influence model predictions. Techniques such as Class Activation Mapping (CAM) highlight influential regions by combining activation maps with output weights~\cite{zhou2016learning}. Layer-wise Relevance Propagation (LRP) propagates relevance scores backward through the network to the input space~\cite{bach2015pixel}. Additionally, Concept Activation Vectors (CAVs) quantify how well human-understandable concepts are represented within learned features~\cite{kim2018interpretability}.

    \textit{Gradient-based} methods use gradients of the output with respect to the input to estimate feature contributions. Integrated Gradients (IG) is a gradient-based explainability approach widely used across diverse applications to attribute a model’s prediction to its input features \cite{shi2024ircan}. IG, for instance, averages gradients computed along a path from a baseline input to the actual input~\cite{sundararajan2017axiomatic}. Integrated Hessians build on Integrated Gradients by characterizing pairwise feature interactions in deep networks under an axiomatic attribution framework \cite{janizek2021explaining}. Gradient$\times$Input combines input values with their corresponding gradients to enhance interpretability~\cite{shrikumar2017learning}. FullGrad extends these methods by incorporating both input and bias gradients~\cite{srinivas2019full}, while Guided IG introduces layer-specific propagation rules to improve attribution clarity~\cite{kapishnikov2021guided}.

\textit{Attention-based} explainability techniques take advantage of attention mechanisms in models like Transformers to visualize and quantify input relevance. Attention Flow and Attention Rollout propagate attention scores through layers to trace token influence~\cite{zhu2018hierarchical, abnar2020quantifying}. These approaches help visualize the pathways through which information flows in deep networks~\cite{vig2019multiscale}.

\textit{Hybrid Methods} improve faithfulness and by combining multiple explanation signals. Examples include combining gradient and activation maps~\cite{selvaraju2020grad}, blending perturbation with activation or attention~\cite{chefer2021transformer, shrikumar2017learning}, and integrating gradients with attention mechanisms~\cite{qiang2022attcat, yuan2021explaining}. AttnLRP~\cite{achtibat2024attnlrp, chefer2021transformer}, which extends Layer-wise Relevance Propagation to Transformer models by explicitly incorporating attention into relevance propagation. These fusion approaches are motivated by the complementary strengths of different explanation sources~\cite{mersha2025explainable, mersha2024explainable}.

\subsection{Explainability for Transformer-based Models}

Transformer-based architectures introduce substantial challenges for explainability due to their depth, self-attention mechanisms, and distributed contextual representations. Perturbation-based methods, such as SHAP~\cite{lundberg2017unified} and LIME~\cite{ribeiro2016should}, while intuitive and model-agnostic, are computationally expensive and often generate inconsistent attributions when applied to models with deep and non-linear architectures~\cite{fantozzi2024explainability, serrano2019attention}. Their sampling-based approximations also disrupt the underlying dependencies within token sequences, leading to misleading explanations in Transformer-based models.

Activation-based techniques, including Class Activation Mapping (CAM)~\cite{zhou2016learning}, LRP~\cite{bach2015pixel}, and Concept Activation Vectors (CAVs)~\cite{kim2018interpretability}, can highlight neuron responses but struggle to provide semantically grounded, fine-grained attributions. This limitation arises from their inability to isolate feature-level interactions or propagate relevance through architectural components like self-attention, residual connections, and normalization layers~\cite{chen2024interpretable}.

Gradient-based methods—including Integrated Gradients (IG)~\cite{sundararajan2017axiomatic} and SmoothGrad~\cite{smilkov2017smoothgrad}—are efficient and theoretically sound. However, these methods typically generate token-level attributions at a final layer without capturing the contextual evolution of token semantics across the model hierarchy. As a result, they fail to faithfully explain how different layers contribute to the final decision, particularly in deeper Transformer stacks where abstract and distributed semantics emerge~\cite{jain2023inseq}.

Attention-based explanations, such as attention weights, Attention Flow, and Attention Rollout~\cite{abnar2020quantifying, ferrando2024primer}, focus on the model’s internal alignment patterns but are limited in scope. They typically emphasize query-key interactions and ignore the contributions of value vectors, feedforward sublayers, residual connections, and LayerNorm operations—components that are central to the model’s reasoning pipeline~\cite{chefer2021transformer}.

Recent Transformer explainability studies have moved beyond raw attention visualization. Flow-based approaches, such as Generalized Attention Flow~\cite{azarkhalili2025generalized}, formulate attribution as a structured flow over attention and gradient signals to better capture cross-layer information routing. AttnLRP~\cite{achtibat2024attnlrp} extends Layer-wise Relevance Propagation to Transformers by incorporating attention mechanisms into relevance redistribution. Complementarily, Contrast-CAT~\cite{han2025contrast} introduces a contrastive activation-based paradigm that suppresses class-irrelevant features to yield sharper token-level explanations.
Furthermore, numerous studies have shown that attention weights alone are insufficient for reliable explanations~\cite{jain2019attention, wiegreffe2019attention}, as they do not necessarily correlate with a model’s actual decision-making process.

To address these limitations, recent efforts have explored relevance propagation extensions. Ali et al.~\cite{ali2022xai} proposed \textit{Conservative Propagation}, combining LRP and Gradient$\times$Input to enhance stability across LayerNorm and attention heads. Chefer et al.~\cite{chefer2021transformer} introduced a Taylor-decomposition-inspired relevance flow that integrates residual and feedforward components. Hou et al.~\cite{hou2023decoding} proposed \textit{Decoding Layer Saliency}, identifying saliency-rich Transformer layers but without attributing contextual evolution of relevance across the entire network.

Despite these advances, a persistent gap remains: existing techniques either aggregate relevance globally, losing layer-specific nuances, or fail to account for contextual shifts that emerge through hierarchical processing. To overcome this, we propose a hierarchical \textbf{Context-Aware Layer-wise Integrated Gradient} (CA-LIG) framework. Our approach decomposes relevance contributions across each Transformer block, while maintaining alignment with the evolution of contextual tokens. Specifically, we compute intermediate relevance scores at each layer, normalize them within context windows, and integrate them with directional signals from gradient-weighted attention scores.

By combining hierarchical relevance decomposition, context-sensitive normalization, and tunable relevance aggregation, our CA-LIG framework advances the state of the art in Transformer explainability, providing more granular, interpretable explanations that address the limitations of existing XAI techniques.

\section{Methodology} \label{sec:method}

\begin{figure*}[ht]
\centering
\includegraphics[width=\textwidth]{ 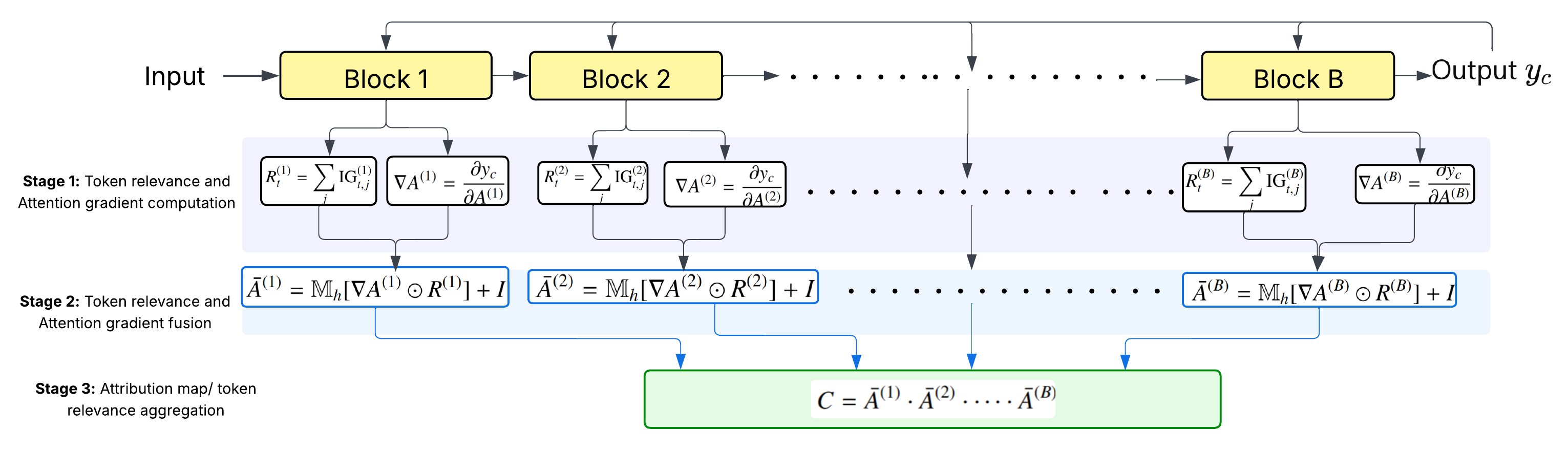}
\caption{Proposed architecture of the Context-Aware Layer-wise Integrated Gradients (CA-LIG) framework.
}
\label{Fig:Architecture}
\end{figure*}

The CA-LIG framework goes beyond final-layer attribution. It consists of four tightly coupled stages. First, \emph{Layer-wise Integrated Gradients} are computed at each Transformer block to quantify how intermediate hidden representations contribute to the model's prediction. Second, \emph{class-specific attention gradients} are derived to capture how information flows through attention heads and how token interactions influence the output. Third, these two signals are \emph{fused through a context-aware integration mechanism}, producing unified relevance scores that reflect both local token importance and global structural dependencies. Finally, a \emph{context-aware attribution step} stabilizes the scores and relevance conservation by rolling relevance across layers to produce the final signed attribution map. Figure~\ref{Fig:Architecture} provides an overview of the CA-LIG framework. This design enables CA-LIG to capture how evidence forms, evolves, and interacts across the hierarchy of a Transformer layer, yielding faithful, interpretable, and context-sensitive explanations. The following subsections detail each component of the framework.

\subsection{Layer-wise Token Relevance via Integrated Gradients}
Let's consider a classifier model $M$ with $C$ output classes and a target class $c \in \{1, \dots, C\}$ for which an explanation is desired. The target class $c$ does not need to correspond to the model’s top prediction; the explanation framework allows generating interpretations for any class of interest. 

The transformer architecture comprises a stack of $N$ layers or blocks, where $\textbf{x}^{(n)}$ denotes the input to the $n^{\text{th}}$ layer $L^{(n)}$, for $n = 1, \dots, N$.  $\textbf{x}^{(1)}$ represents the input embeddings of the model, while $\textbf{x}^{(N)}$ corresponds to the output of the final encoder layer. 

Relevance and gradient signals are computed with respect to the target class score $y_c$, allowing the tracking of class-specific evidence throughout the model hierarchy.

We employ a Layer-wise Integrated Gradients (LIG) approach to compute token-level attributions in a faithful and interpretable manner. Rather than attribution of feature importance only at the classifier layer, we extend the IG approach to each intermediate layer. For a given layer $l$, we extract the contextual hidden representation $\textbf{x}^{(l)} \in \mathbb{R}^{s \times d}$, where $s$ denotes the sequence length and $d$ is the hidden dimension size. A corresponding baseline representation $x'^{(l)} \in \mathbb{R}^{s \times d}$ is defined and serves as a neutral reference point.

We construct a trajectory of interpolated hidden states between a baseline and the actual representation of the model for the input to approximate the path integral used in IG. For a given layer $l$, let $x^{(l)} \in \mathbb{R}^{s \times d}$ denote the actual hidden representation of the input sequence and $x'^{(l)} \in \mathbb{R}^{s \times d}$ be the corresponding baseline. We then compute a sequence of interpolated hidden states using the following equation:
\begin{equation}
  x^{(l)}(\alpha_k) = x'^{(l)} + \alpha_k \cdot \left(x^{(l)} - x'^{(l)}\right), \quad \text{where } \alpha_k = \frac{k}{m}, \quad k = 1, \dots, m.  
\end{equation}
Here $m$ is the number of interpolation steps, and $\alpha_k$ defines the step size along the path from baseline to actual representation. 

We compute the gradient of the output score $y_c$ with respect to each interpolated hidden state and then aggregate these gradients to compute the integrated gradients at layer $l$ using a Riemann sum over $m$ steps \cite{sundararajan2017axiomatic}:

\begin{equation}
   \text{IG}^{(l)} = (x^{(l)} - x'^{(l)}) \odot \frac{1}{m} \sum_{k=1}^{m} \frac{\partial y_c}{\partial x^{(l)}(\alpha_k)}, 
\end{equation}
where $\odot$ denotes element-wise multiplication.

We aggregate the feature-wise attributions across the hidden dimensions for each token to obtain token-level relevance scores. For token $t \in \{1, \dots, s\}$, the relevance score at layer $l$ is given by:
\begin{equation} \label{Token Relevance}
    R_t^{(l)} = \sum_{j=1}^{d} \text{IG}_{t,j}^{(l)},
\end{equation} 
where\( j \in \{1, \dots, d\} \) is the hidden dimension indices of a token representation.
The use of this summation is grounded in attribution theory \cite{sundararajan2017axiomatic}. The Integrated Gradients method satisfies the completeness property, which ensures that the sum of all attributions equals the difference in the model’s output between the input and the baseline. Summing the feature-wise attributions for each token preserves this property and provides a faithful measure of the total influence that a token exerts on the model’s decision.

The result is a signed, layer-wise attribution map, where positive values indicate supportive evidence and negative values indicate opposing influence on the target class decision.

\subsection{Compute  the Gradient of Attention per Transformer Block}

Based on token-level relevance scores obtained through LIG, we now examine how token-to-token contextual interactions, captured by the self-attention mechanism, contribute to the model's decision for a given target class $y_c$. This approach helps to move beyond the importance of isolated tokens and toward understanding how the model leverages structured dependencies between tokens to form and make its prediction.

Each transformer block $b$ contains a self-attention mechanism that produces an attention matrix $A^{(b)} \in \mathbb{R}^{h \times s \times s}$, where $h$ is the number of attention heads and $s$ is the sequence length. The attention score $A_{i,j,k}^{(b)}$ represents how much attention token $j$ pays to token $k$ in the head $i$ of block $b$, computed in \ref{attention_weight}.
\begin{equation} \label{attention_weight}
  A^{(b)} = \text{softmax} \left( \frac{Q^{(b)} (K^{(b)})^\top}{\sqrt{d_k}} \right),  
\end{equation}

where $Q^{(b)}$ and $K^{(b)}$ are the queries and keys at block $b$, respectively, computed from the input to the block $X^{(b)}$, and $d_k$ denotes the dimensionality of the key vectors \cite{vaswani2017attention}.

To quantify the class-specific influence of these attention weights, we compute the gradient of the output score $y_c$ with respect to the attention matrix using equation \ref{Attention gradients}.
\begin{equation} \label{Attention gradients}
    \nabla A^{(b)} = \frac{\partial y_c}{\partial A^{(b)}} \in \mathbb{R}^{h \times s \times s}.
\end{equation}

The $\nabla A^{(b)}$ gradient tensor captures the sensitivity of the prediction of the model to changes in individual attention weights \cite{chefer2021transformer}. A high magnitude in $\nabla A_{i,j,k}^{(b)}$ implies that the connection of the attention of the token $j$ to the token $k$ in the head $i$ has a strong influence on the output of the class $y_c$. These gradients provide a class-specific saliency map of the attention structure and allow us to identify token interactions in the input sequence that contribute to the model's reasoning. By interpreting this sensitivity map, we reveal the structural dependencies on which the model relies, highlighting not only which tokens are important but also how they interact and influence each other through attention mechanisms.

\subsection{Layer-wise Relevance and Attention Gradient Fusion}

We combine token-level relevance scores obtained from LIG through Equation \ref{Token Relevance} with attention-based sensitivity maps computed from Equation \ref{Attention gradients} to create a \textit{context-aware explanation}. This integration allows for capturing \emph{local importance} (the individual contribution of each token) and \emph{contextual influence} (how the relationships between tokens affect the model's prediction).

For each transformer block $b$, we perform an element-wise combination of the attention gradients $\nabla A^{(b)} \in \mathbb{R}^{h \times s \times s}$ with a normalized form of token-level relevance $\text{Norm}(R^{(l)}) \in \mathbb{R}^{s}$, where $R^{(l)}$ is the relevance vector from layer $l$, which aligned to the attention map’s sequence dimension. We apply the Symmetric Min-Max Normalization method.  The fusion is defined as:

\begin{equation} \label{Token relevance and Attention gradient fusion}
    R_{\text{context}}^{(b)} = \nabla A^{(b)} \odot \text{Norm}(R^{(l)}),
\end{equation}
where $\odot$ denotes the Hadamard (element-wise) product.
This operation weights the attention gradients by the relative importance of each token, thereby enriching the attribution with contextual interactions while preserving the fidelity of individual token contributions.
The element-wise fusion in Equation~\ref{Token relevance and Attention gradient fusion} and ~\ref{tunable_attention_fusion} operates as a relevance-gated sensitivity mechanism that ensures faithfulness in attention–relevance fusion. Token-level relevance scores obtained via Layer-wise Integrated Gradients provide causally grounded attributions that indicate whether a token contributes to the model’s prediction, while attention gradients capture how sensitive the prediction is to inter-token attention pathways. This design preserves token-level attribution faithfulness while extending explanations to capture context-aware, interaction-level effects within transformer representations.

\subsection{Context-Aware Attribution via Attention-Relevance Fusion}
To obtain a unified and context-aware attribution map, we aggregate relevance information across attention heads and transformer layers. This aggregation captures both the sensitivity of attention pathways and the contextual contribution of input tokens. To achieve this, we introduce a tunable fusion coefficient \( \lambda \in [0, 1] \) that balances two complementary gradient-based relevance signals. Specifically, for each transformer block \( b \in \{1, \dots, B\} \), where \( B \) is the total number of blocks, we compute a fused attention relevance matrix as follows:

\begin{equation} \label{tunable_attention_fusion}
  \bar{A}^{(b)} = \mathbb{M}_h \left[ \lambda \cdot \left( \nabla A^{(b)} \odot \text{Norm}(R^{(l)}) \right) + (1 - \lambda) \cdot \text{Norm}(R^{(l)}) \right] + I,
\end{equation}
where, \( \nabla A^{(b)} \in \mathbb{R}^{h \times s \times s} \) denotes the gradient of attention weights at block \( b \) with respect to the target score \( y_c \), and \( R^{(l)} \in \mathbb{R}^s \) represents token-level relevance, normalized as \( \text{Norm}(R^{(l)}) \), \( \odot \) denotes the element-wise product. The averaging operator \( \mathbb{M}_h[\cdot] = \frac{1}{h} \sum_{i=1}^h (\cdot) \) aggregates across heads, and the identity matrix \( I \in \mathbb{R}^{s \times s} \) preserves residual connections.

The main advantage of introducing \( \lambda \) lies in its ability to modulate the degree of influence from the sensitivity of the attention weights and the relevance of the input level. When \( \lambda = 1 \), the explanation balances attention-gradient information and token-level relevance. In contrast, \( \lambda = 0 \), the explanation is purely driven by the input token.
This controlled attribution method is useful for transformer architectures, where syntax and semantics are distributed across layers and heads.

Each fused matrix \( \bar{A}^{(b)} \) is row-normalized before composition to maintain bounded influence scores. To trace the flow of information from input to deeper layers, we recursively multiply the normalized relevance-weighted attention matrices across blocks \cite{abnar2020quantifying, chefer2021transformer}. 

\begin{equation} \label{rollout_composition}
    C = \bar{A}^{(1)} \cdot \bar{A}^{(2)} \cdot \dots \cdot \bar{A}^{(B)},
\end{equation}
The resulting matrix \( C \in \mathbb{R}^{s \times s} \) constitutes the final context-aware attribution map, where each entry \( C_{i,j} \) indicates the cumulative influence of token \( j \) on the representation of token \( i \).

This relevance map can be decomposed into positive and negative components to interpret supporting versus opposing contributions to the target prediction.

\begin{equation}
\label{Equ:positive vs negative attribution}
    C^+ = \max(0, C), \quad C^- = \min(0, C),
\end{equation}

where $C^+$ reflects supportive evidence and $C^-$ indicates inhibitory influence on the predicted class. This decomposition provides fine-grained insight into how different parts of the input promote or counteract the model’s decision.

\subsection{Methodological Comparison of CA-LIG with Prior Methods}
To understand how CA-LIG differs from existing attribution techniques, a clear methodological comparison across key aspects is helpful. Prior methods such as Integrated Gradients~\cite{sundararajan2017axiomatic}, Integrated Hessians~\cite{janizek2021explaining}, and Transformer-LRP ~\cite{chefer2021transformer} each capture different aspects of model reasoning. In contrast, CA-LIG integrates layer-wise Integrated Gradients with class-specific attention-gradient fusion, enabling hierarchical and context-aware explanations that track how evidence evolves across Transformer layers. Table~\ref{tab:methodlogicalComparison} in ~\ref {appendix A} provides a consolidated comparison of these methods, highlighting the distinctions that motivate the design of the CA-LIG framework.

\section{Experimental Setups} \label{sec:experiments}
To evaluate the effectiveness of our proposed approach, we conducted experiments on multiple encoder-based Transformer language models trained on diverse NLP tasks and datasets. We also evaluated it on the vision task for the domain reproducibility of our approach.
We benchmark our approach against widely used explainability techniques. All experiments were conducted on Google Colab using an NVIDIA A100-SXM4 GPU (40GB VRAM), CUDA 12.2, PyTorch 2.0.1, and HuggingFace Transformers 4.35.

Integrated Gradients is computed using a linear interpolation path between a zero embedding baseline and the input embedding. We use 50 interpolation steps because they provide sufficient numerical accuracy, stable attributions, and balanced computational efficiency. Token-level relevance scores are normalized using L1 normalization within each layer. Attention rollout depth includes all encoder layers, and attention gradients are aggregated via mean pooling across heads. Sensitivity analysis over 25, 50, and 100 steps showed stable attribution trends.  For all experiments, random seeds were fixed to ensure reproducibility, and models were fine-tuned. Fine-tuning was performed for a fixed number of epochs under identical training conditions across all compared XAI methods to ensure fair evaluation. 

The total data scale is detailed in Section 4.1. Each experimental evaluation uses 20\% of each dataset as the held-out test set, with the remaining data allocated to training. To ensure reproducibility, stability, and reliability, all experiments were conducted using independent repeated runs. For each dataset–model configuration, experiments were repeated 10 times under fixed random seeds and identical training conditions. Attribution maps were generated independently in each run, enabling a controlled, systematic evaluation of explanation consistency and robustness across repetitions rather than relying on isolated single-run outcomes.

\subsection{Datasets}
We employ well-known NLP and vision datasets to ensure a comprehensive evaluation of our method. In NLP, we use the IMDB movie reviews dataset, which contains 50,000 reviews (25,000 for training and 25,000 for testing) for binary sentiment classification \cite{maas2011learning}. We also include the Amharic hate speech dataset, an under-resourced language resource consisting of 15,100 annotated samples for hate speech classification \cite{ayele2023exploring}. To further examine the model’s ability to capture long-range contextual dependencies, we incorporate the 20 Newsgroups dataset \cite{lang1995newsweeder}, a collection of long-form documents containing 20 categories.

In vision, we used the CIFAR-10 dataset, comprising 60,000 images in 10 classes \cite{krizhevsky2009learning}. We also include the ASIRRA dataset (cat vs dog) \cite{elson2007asirra} for additional evaluation of the quality of the explanation, as its class-defining features are clear and visually interpretable.

\subsection{Models}
For text classification, we use BERT-base \cite{devlin2019bert}, which processes sequences of up to 512 tokens, with the \texttt{[CLS]} token passed to a classification head for label prediction. To address low-resource settings, we additionally employ XLM-RoBERTa (XLM-R) \cite{conneau2019unsupervised} and AfroLM \cite{dossou2022afrolm}, a multilingual model effective across under-resourced languages. AfroLM is a self-active-learning multilingual language model for 23 African languages. In the visual task, we use a Masked Autoencoder (MAE) \cite{he2022masked}.

\subsection{Baseline XAI Methods}

We compare our proposed approach with well-established attribution techniques commonly used to interpret Transformer-based encoder models. Specifically, we benchmark our method against Input × Gradient (IxG) \cite{shrikumar2017learning}, Integrated Gradients \cite{sundararajan2017axiomatic}, and Layer-wise Relevance Propagation (LRP) \cite{bach2015pixel}, which represent two primary explanation paradigms: gradient-based and relevance propagation-based. We also include Attention Rollout \cite{abnar2020quantifying} and Attention Last (Attention-last) \cite{hollenstein2021relative, sood2020interpreting} to capture cumulative attention flow across layers, providing a complementary perspective to gradient-based attributions.

Our proposed approach can be instantiated in multiple variants to explore the effects of contextual and hierarchical attribution across Transformer layers. We experimented with the following three configurations. First, Context-Aware Integrated Gradients (CA-IG) (at last layer), ($\bar{A}^{(B)}$
), applies at the final layer, attributing relevance based on the output representation of the final transformer layer. Second, CA-IG (Layered) computes attributions independently at each layer, providing a detailed layer-wise view of token relevance and shedding light on how contextual representations evolve throughout the layers. Third, \textbf{CA-LIG} aggregates token-level attributions across layers by composing relevance scores using a rollout strategy. This variant captures the hierarchical flow of contextual importance by propagating relevance through the model's layered structure while maintaining the contributions of token interactions at each level. Furthermore, our approach supports attribution analysis at selective depths, such as earlier, middle, or deepest layers, highlighting the flexibility and strength of our approach in interpreting layer-wise model behavior.

\subsection{Evaluation Settings}
We adopt a two-fold evaluation strategy for the linguistic domain, combining the ERASER benchmark~\cite{deyoung2019eraser} with the framework proposed by \cite{mersha2025evaluating}. From ERASER, we use the Movie Reviews dataset for binary sentiment classification, which provides human-annotated rationales and thus enables the assessment of explanation quality against gold references. This evaluation allows us to measure not only the alignment of model explanations with human rationales but also their stability and coherence.


\begin{figure*}[t]
\centering

\begin{minipage}{0.87\textwidth}
    \centering
    \includegraphics[width=1\textwidth]{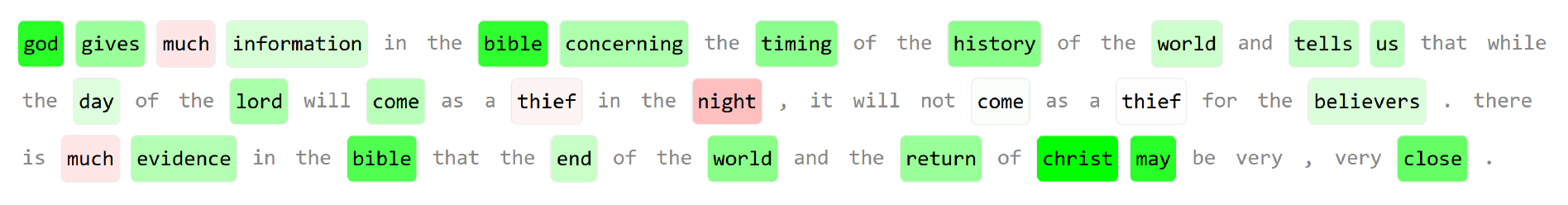} 
    \caption{CA-LIG token-level attributions for a document labeled Christian class from the 20 Newsgroups dataset using BERT-large. Brighter green tokens provide stronger positive evidence, lighter green indicates weaker support, red shows negative influence, and white denotes neutral relevance.}
    \label{Fig:Christian class}
\end{minipage}

\vspace{0.5cm}

\begin{minipage}{0.8\textwidth}
    \centering
    \includegraphics[width=1\textwidth]{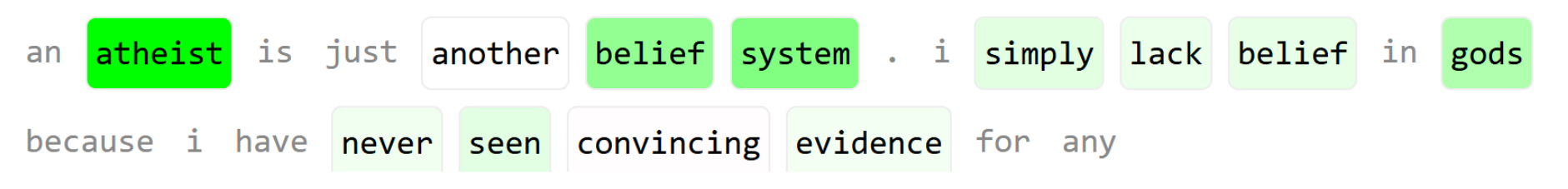} 
    \caption{CA-LIG token-level attributions for a \textit{document labeled atheist class} from the 20 Newsgroups dataset using BERT-base. Brighter green tokens provide stronger positive evidence, lighter green indicates weaker support, red shows negative influence, and white denotes neutral relevance.}
    \label{Fig:atheist class}
\end{minipage}

\vspace{0.5cm}

\begin{minipage}{0.8\textwidth}
    \centering
    \includegraphics[width=1\textwidth]{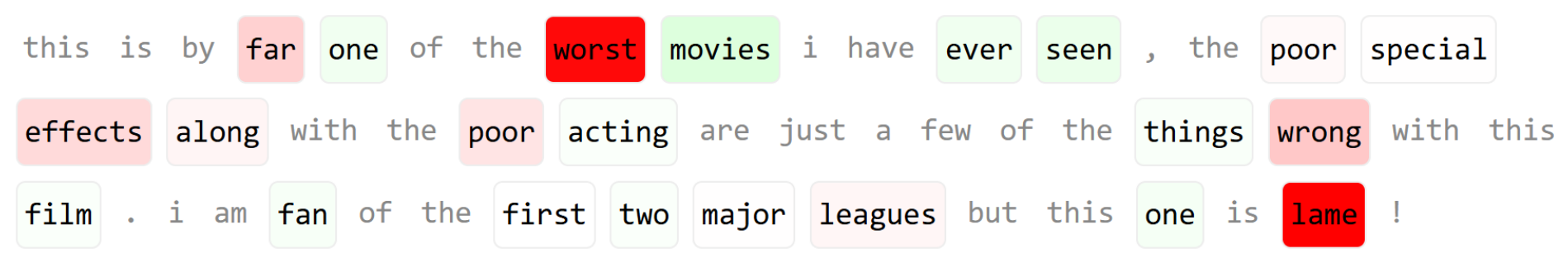} 
    \caption{CA-LIG token-level attributions for a negative IMDB review using BERT-Large. Brighter red indicates stronger negative evidence, green indicates positive relevance, and white denotes neutral tokens.}    
 \label{Fig:Negative Sample2}
\end{minipage}
\end{figure*}


\section{Results} \label{sec:results}
We present the results of the proposed CA-LIG framework, beginning with qualitative visualizations, followed by quantitative comparisons with baseline methods, and concluding with layer-wise attribution case analyses that highlight the effectiveness of our proposed framework across various Transformer models and tasks. All experiments are conducted with $\lambda = 1$, which provides a balanced fusion of attention-gradient information and token-level relevance since our objective is to generate context-aware explanations.

\begin{figure*}[t]
\centering

\begin{minipage}{0.8\textwidth}
    \centering
    \includegraphics[width=0.6\textwidth]{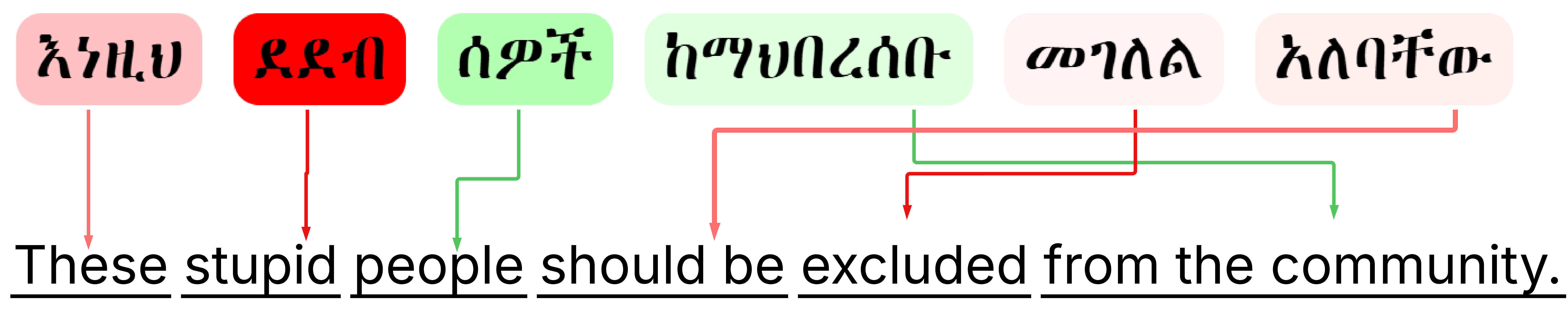} 
    \caption{CA-LIG token-level attributions for an Amharic hate speech sample using the XLM-R model. Brighter red indicates stronger negative evidence, green indicates positive relevance, and white denotes neutral tokens. } 
    \label{Fig:hate Sample}
\end{minipage}

\vspace{0.3cm}

\begin{minipage}{0.8\textwidth}
    \centering
    \includegraphics[width=0.8\textwidth]{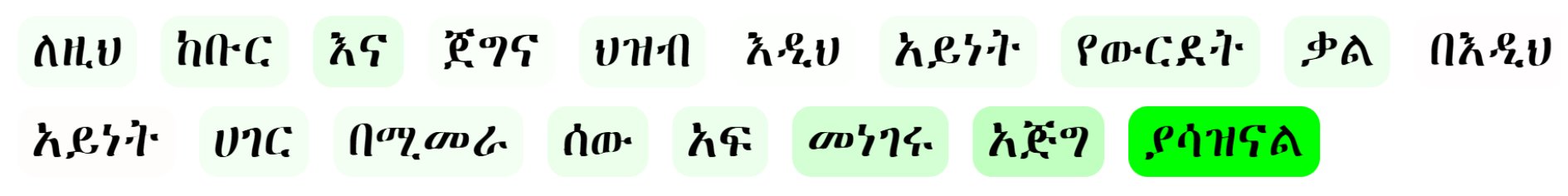} 
    \caption{ CA-LIG token-level attributions for an Amharic not hate speech sample using the AfroLM model. Brighter green indicates stronger positive evidence, red indicates negative relevance, and white denotes neutral tokens.}
    \label{Fig:not hate Sample2}
\end{minipage}
\end{figure*}

We examine the explanations generated by CA-LIG and the baseline methods for various tasks, datasets, models, and domains.
Figures~\ref{Fig:Christian class} and~\ref{Fig:atheist class} present explanations for long-form, multi-class document classification of the 20 Newsgroups dataset using BERT-base and BERT-large models, which demonstrate CA-LIG’s ability to capture long-range contextual dependencies and propagate relevance coherently across long documents. Figure~\ref{Fig:Negative Sample2} presents the negative IMDB sentiment explanation generated by CA-LIG with the BERT-large model. Figures~\ref{Fig:baseline CA-LIG} show token-level relevance patterns for positive IMDB sentiment using BERT-base, comparing CA-LIG with baseline methods. 
Figures~\ref{Fig:hate Sample} and~\ref{Fig:not hate Sample2} evaluate CA-LIG in under-resourced language settings, showcasing Amharic hate-speech detection using XLM-R and AfroLM models. These results confirm that CA-LIG produces stable relevance scores even in morphologically rich, low-resource languages. Figure~\ref{fig:baseline_comparison1} provides explanations generated with CA-LIG using the MAE model, demonstrating the flexibility of CA-LIG beyond text and highlighting how the framework generalizes to vision tasks.

We observe that attention-based methods, Figure \ref{Fig:baseline CA-LIG}, such as A-last and A-Rollout, tend to assign relatively uniform relevance scores across all tokens, with particular amplification of salient tokens like \textit{“amazing”} and \textit{“absolutely”}, yet they provide limited contrast in importance among the remaining tokens. IxG, LRP, and IG provide more selective relevance, with LRP focusing heavily on \textit{“amazing”} and IG assigning moderate negative relevance to \textit {“this”} and \textit{“movie”}. Interestingly, IG also overemphasizes the [CLS] token, which may dilute the interpretability of specific input tokens.
Our proposed methods show clearer attribution patterns. CA-IG (last layer) assigns strong positive relevance to sentiment-bearing words (\textit{“absolutely”} and \textit{“amazing”}) while moderately suppressing less informative tokens. CA-LIG further refines this by progressively aggregating relevance across layers, resulting in sharp, focused attribution on the most sentiment-relevant terms. Unlike baseline methods, our CA-IG variant avoids over-weighting special tokens such as [CLS] by redistributing relevance across contextually interacting tokens, thereby preventing special-token dominance and offering a more intuitive, human-aligned explanation of model behavior.
These qualitative observations support the effectiveness of our context-aware method in isolating semantically important words and capturing deeper attribution flow across the Transformer’s architecture. Furthermore, we visualize the head and layer-wise attribution heatmaps derived from \( C \) to identify patterns of structural and contextual importance, which aid in model interpretation, debugging, and trust calibration. Supplementary experiments are provided in \ref{appendix B}.

\begin{figure}[ht]
\centering
\includegraphics[width=8.5cm]{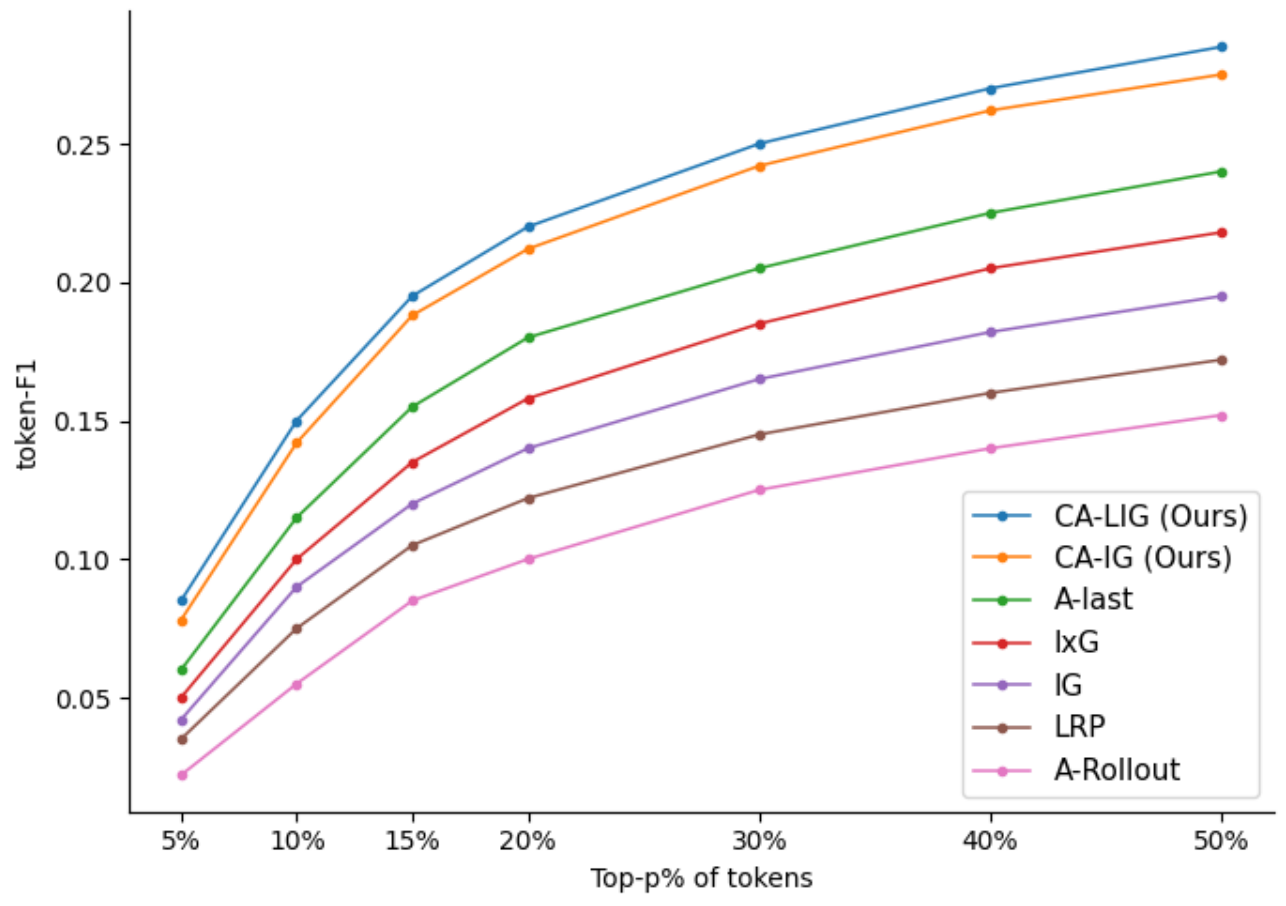}
\caption{Token-F1 scores on the Movie Reviews reasoning task. CA-LIG and CA-IG(last layer) achieve consistently higher performance than baseline methods.
}
\label{Fig:token-F1 evaluation}
\end{figure}

\begin{figure}[ht]
\centering
\includegraphics[width=8.2cm]{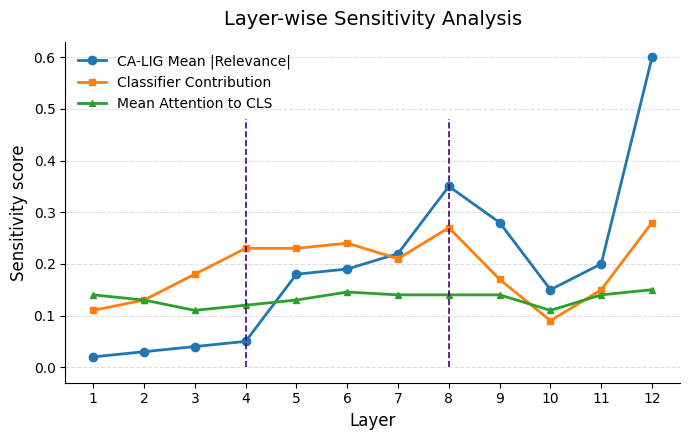}
\caption{Layer-wise sensitivity profiles for CA-LIG, classifier contribution, and mean attention to \texttt{[CLS]}. Transitional peaks at Layers 4 and 8 reflect key representational shifts. The final peak at Layer 12 corresponds to sentiment consolidation.}
\label{Fig:Case Sensitivty Analysis}
\end{figure}

Figure~\ref{fig:baseline_comparison1} shows a qualitative comparison for image classification. Baseline methods often highlight scattered or background regions, which limit interpretability. In contrast, our approach yields more coherent, concentrated visualizations, emphasizing areas directly relevant to the predicted class.

\begin{figure*}[t]
\centering
\setlength{\tabcolsep}{2pt}
\renewcommand{\arraystretch}{0}

\begin{tabular}{cccccc}
\small\textbf{Input} & \small\textbf{Grad-CAM \cite{selvaraju2017grad}} & \small\textbf{LRP-$\epsilon$ \cite{bach2015pixel}} & \small\textbf{LRP(Attn+Grad) \cite{chefer2021transformer}} & \small\textbf{IG \cite{sundararajan2017axiomatic}} & \small\textbf{CA-LIG (Ours)} \\[2pt]

\includegraphics[width=0.13\textwidth]{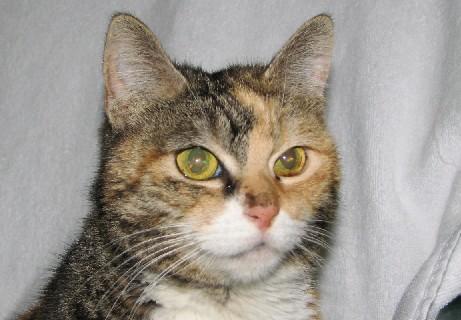} &
\includegraphics[width=0.13\textwidth]{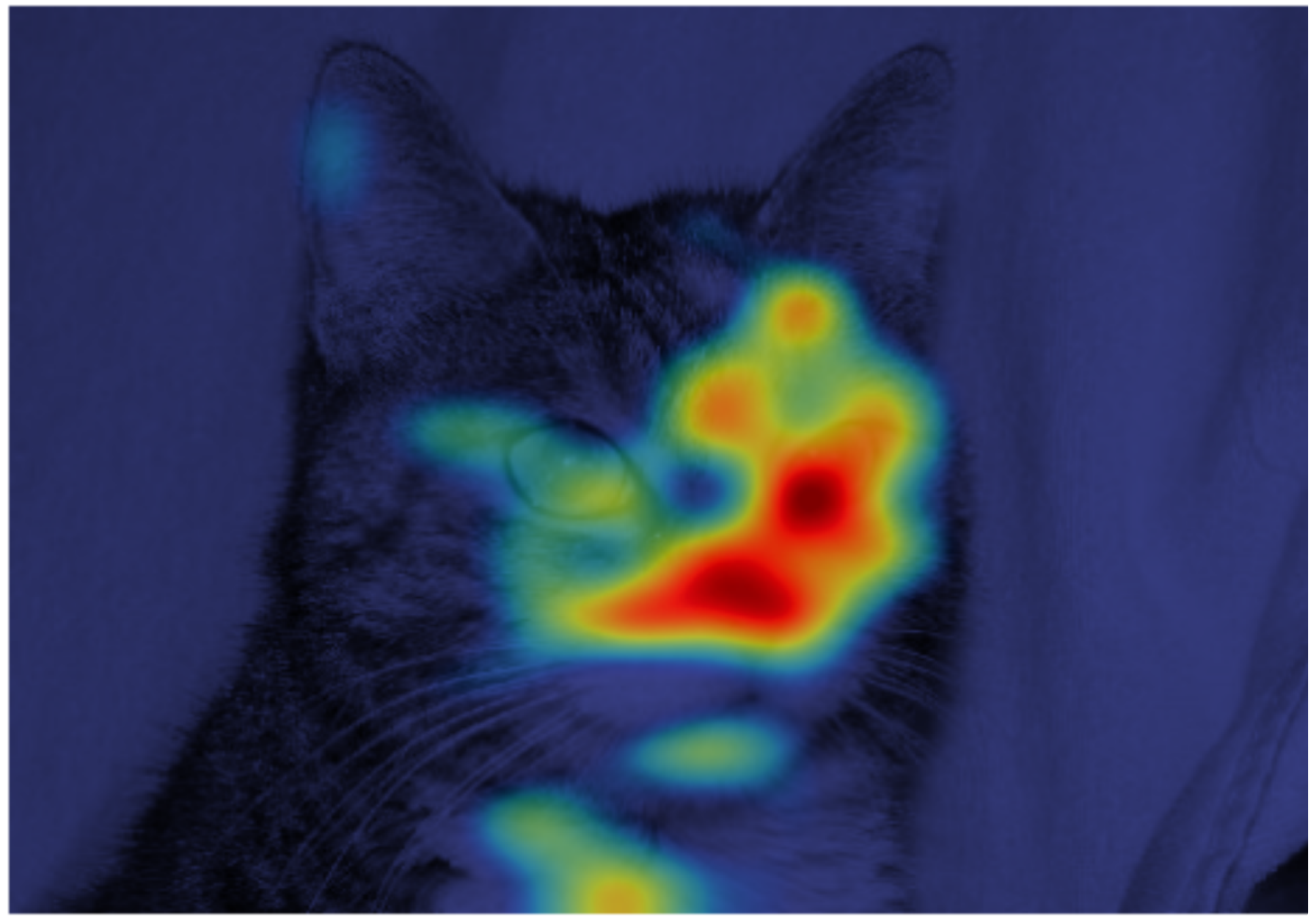} &
\includegraphics[width=0.13\textwidth]{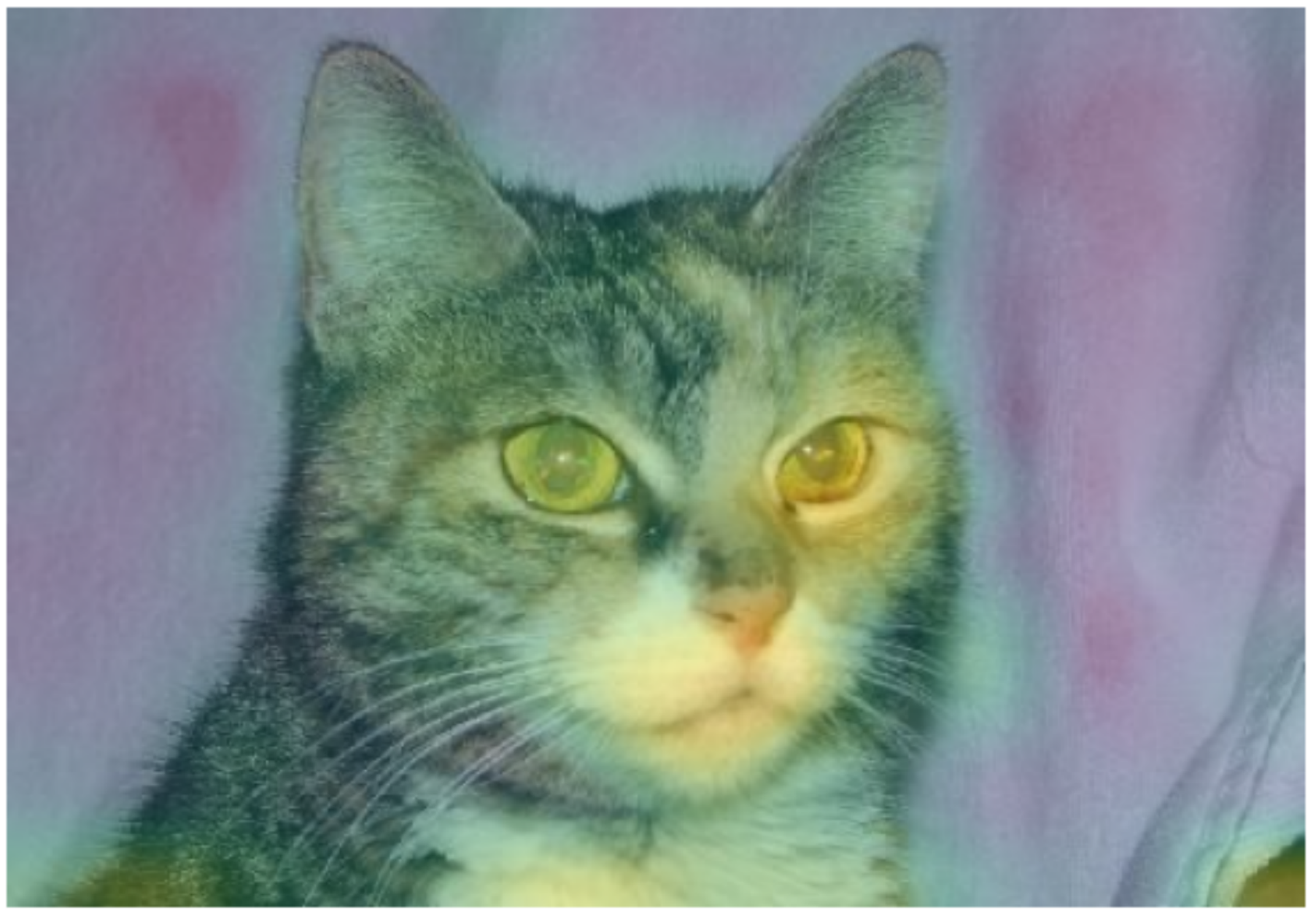} &
\includegraphics[width=0.13\textwidth]{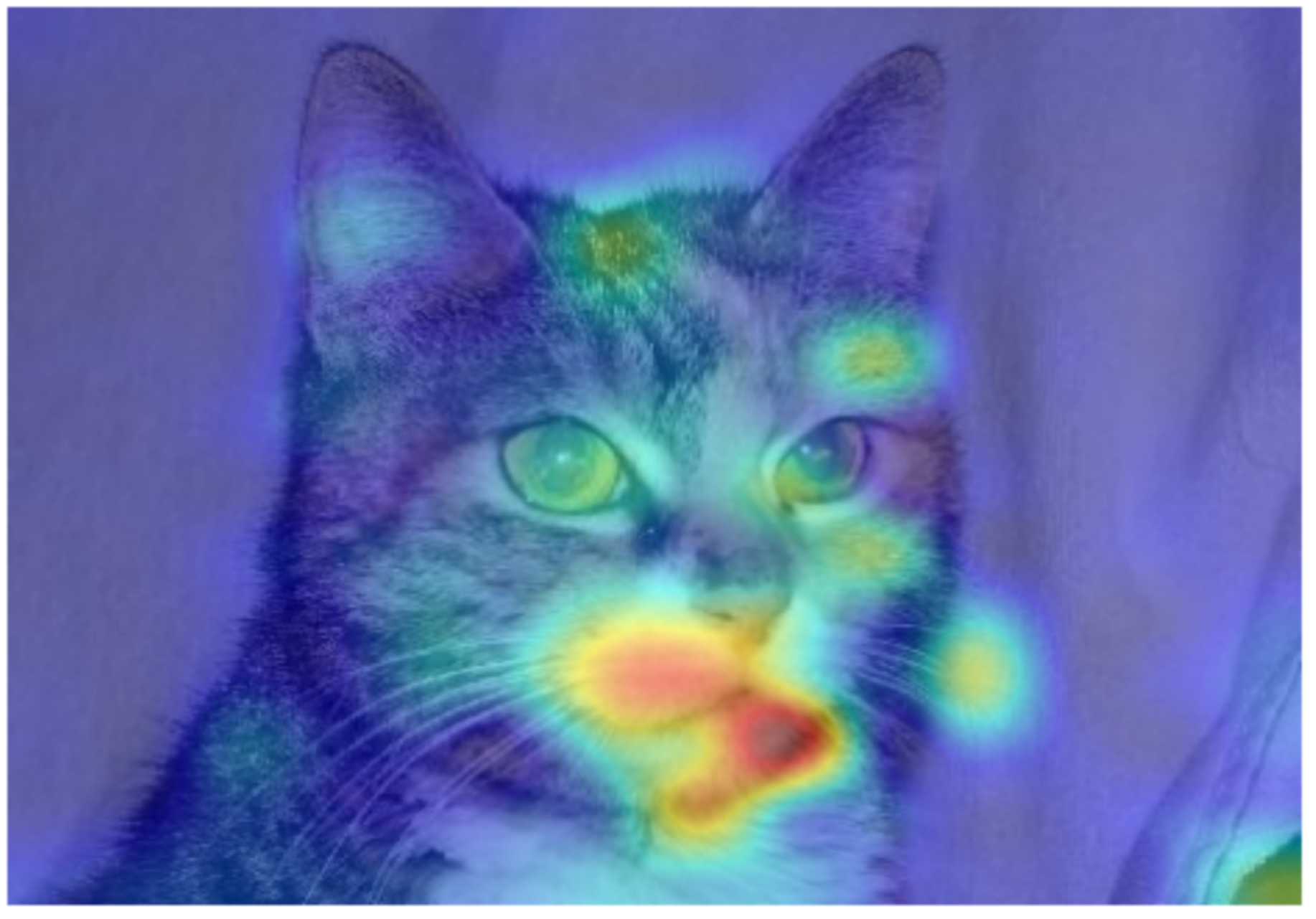} &
\includegraphics[width=0.13\textwidth]{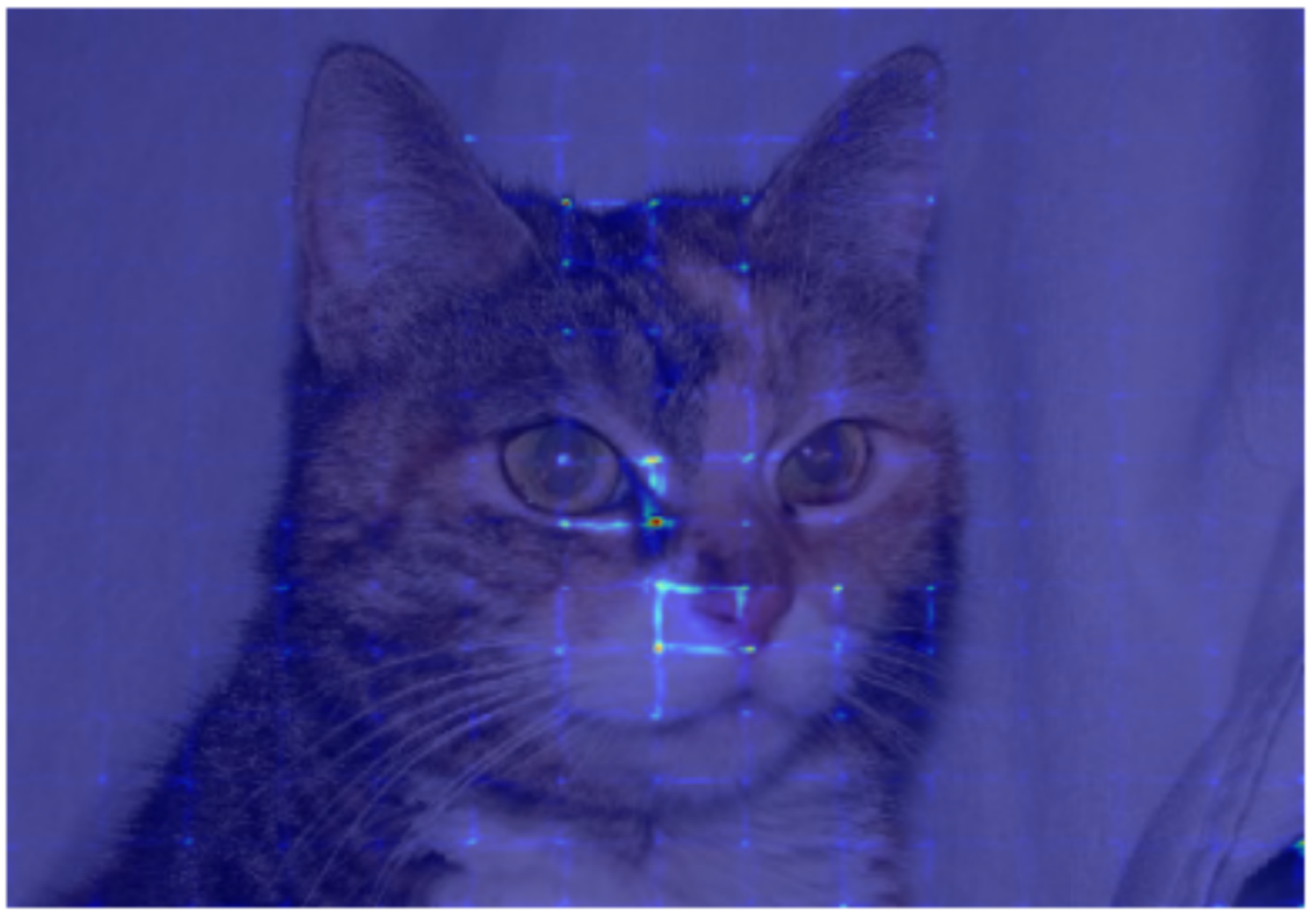} &
\includegraphics[width=0.13\textwidth]{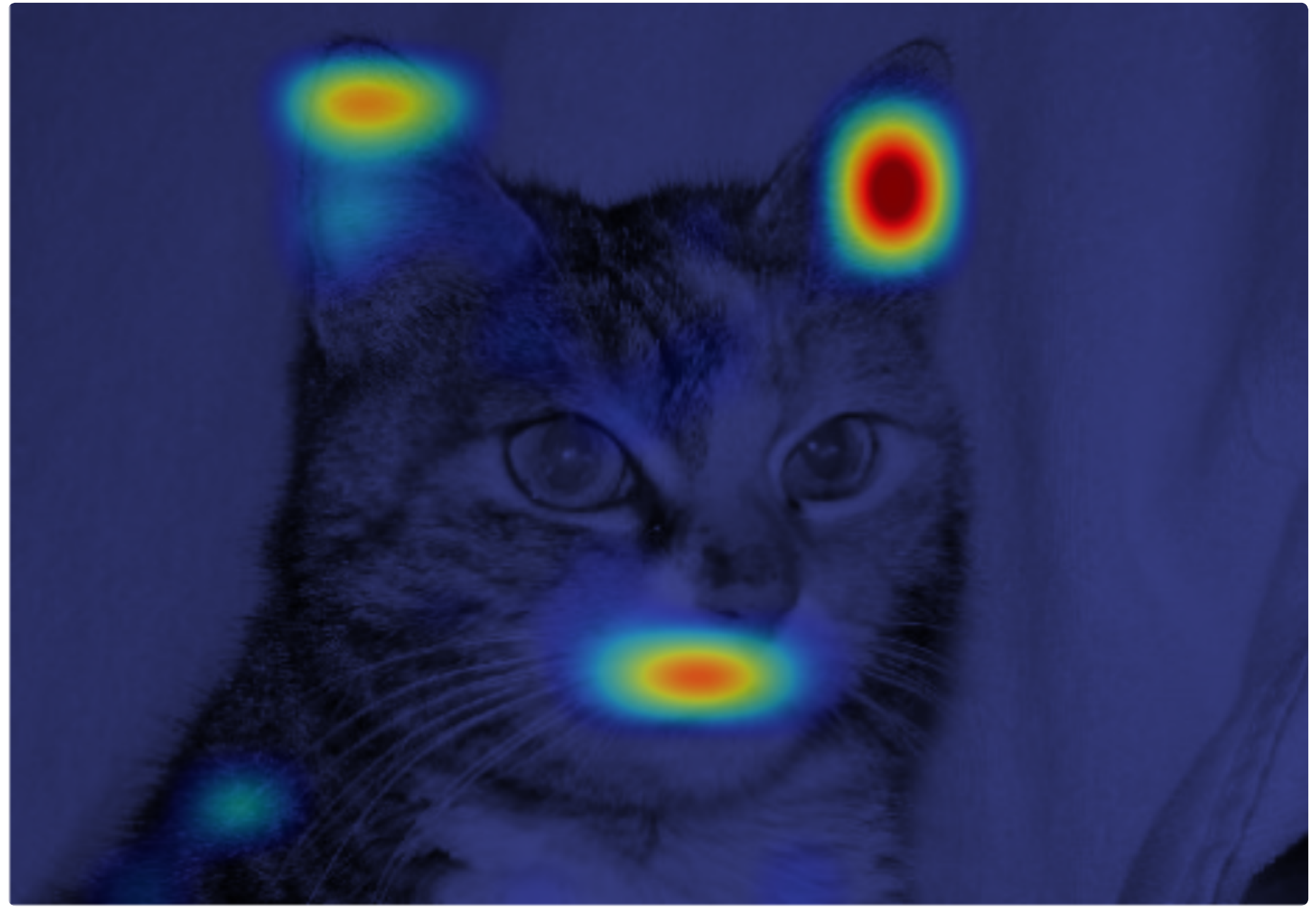} \\[1pt]

\includegraphics[width=0.13\textwidth]{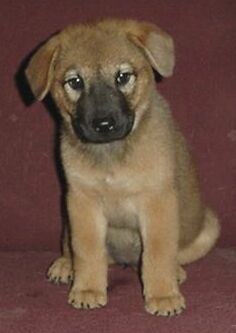} &
\includegraphics[width=0.13\textwidth]{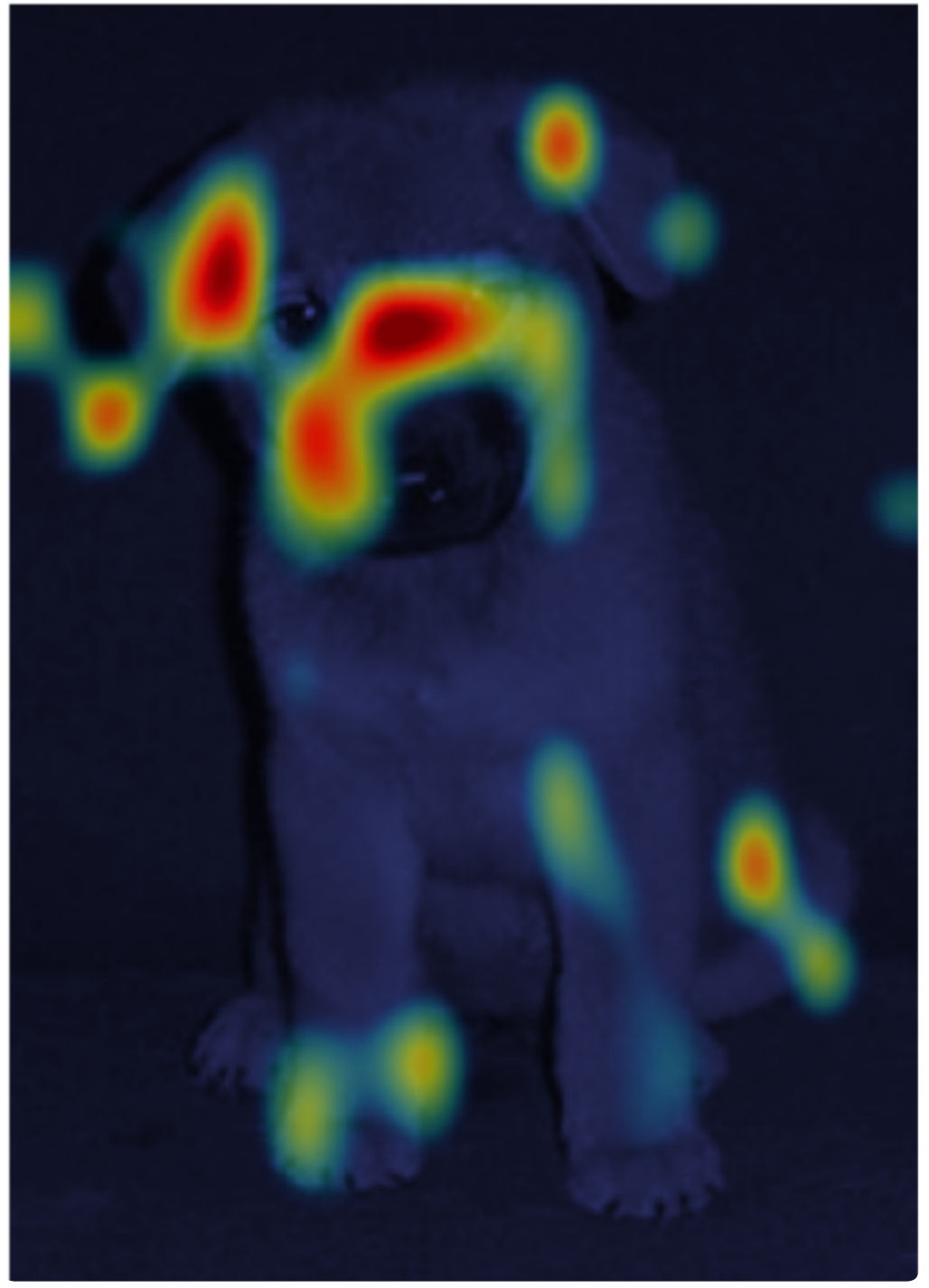} &
\includegraphics[width=0.13\textwidth]{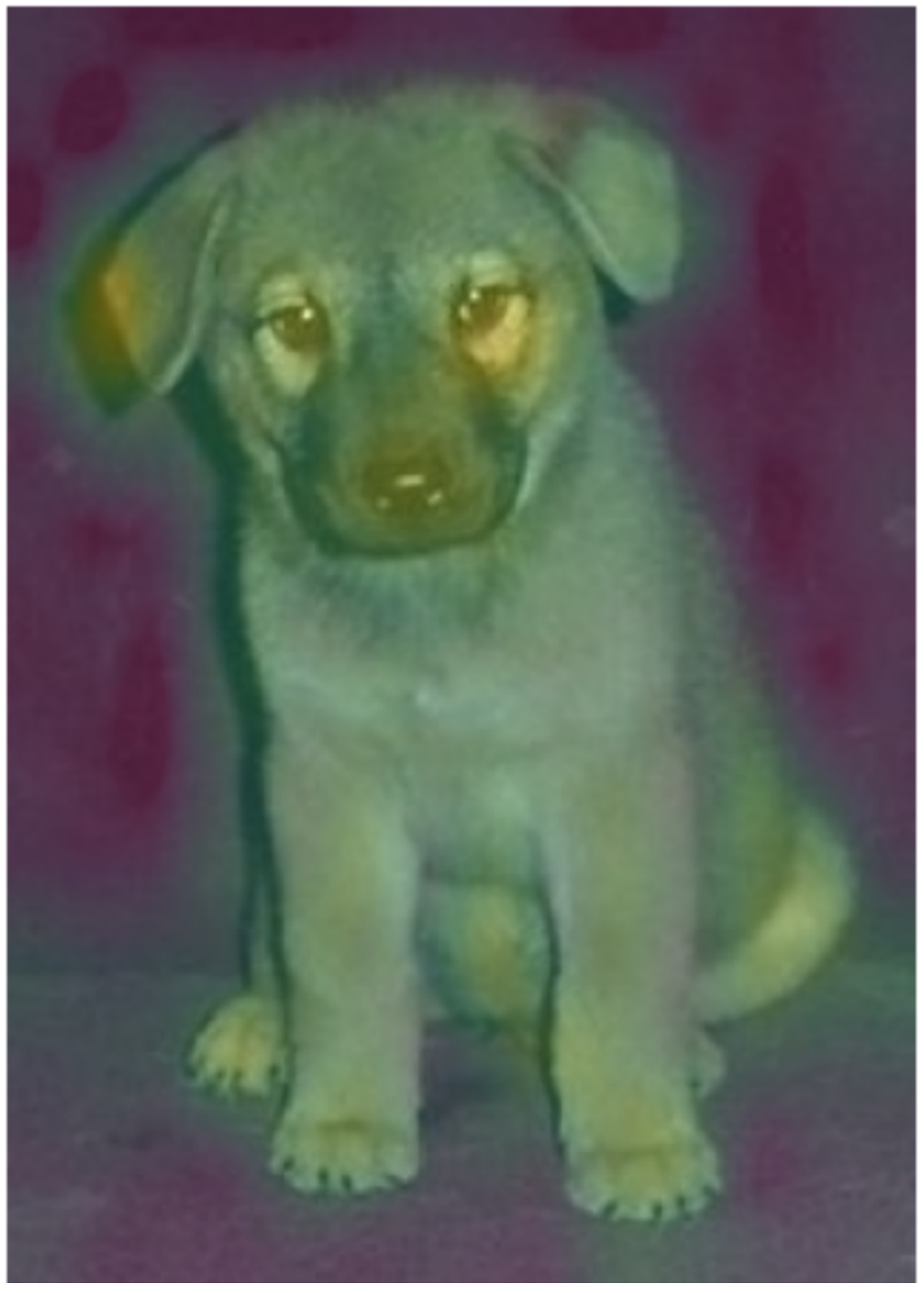} &
\includegraphics[width=0.13\textwidth]{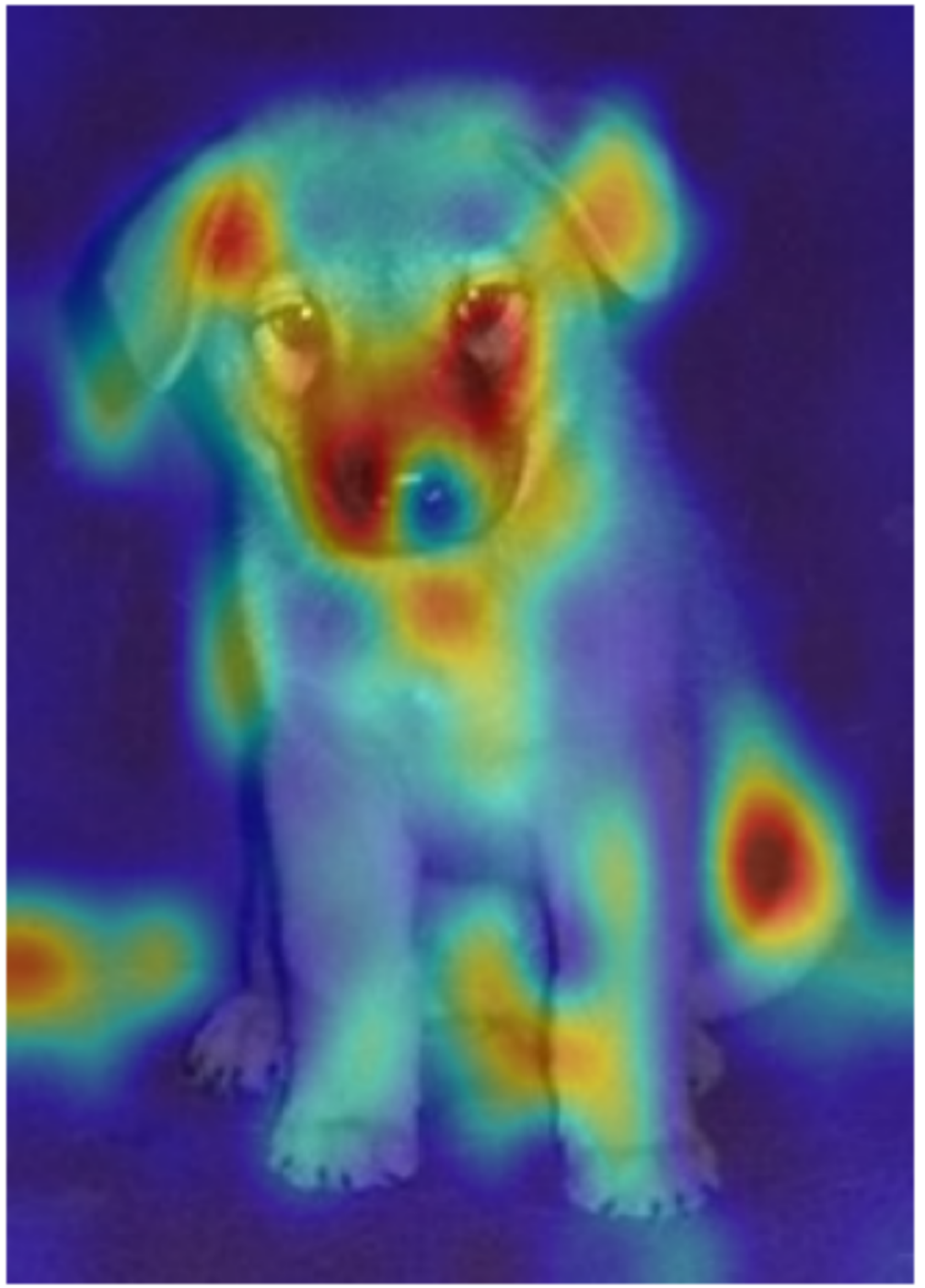} &
\includegraphics[width=0.13\textwidth]{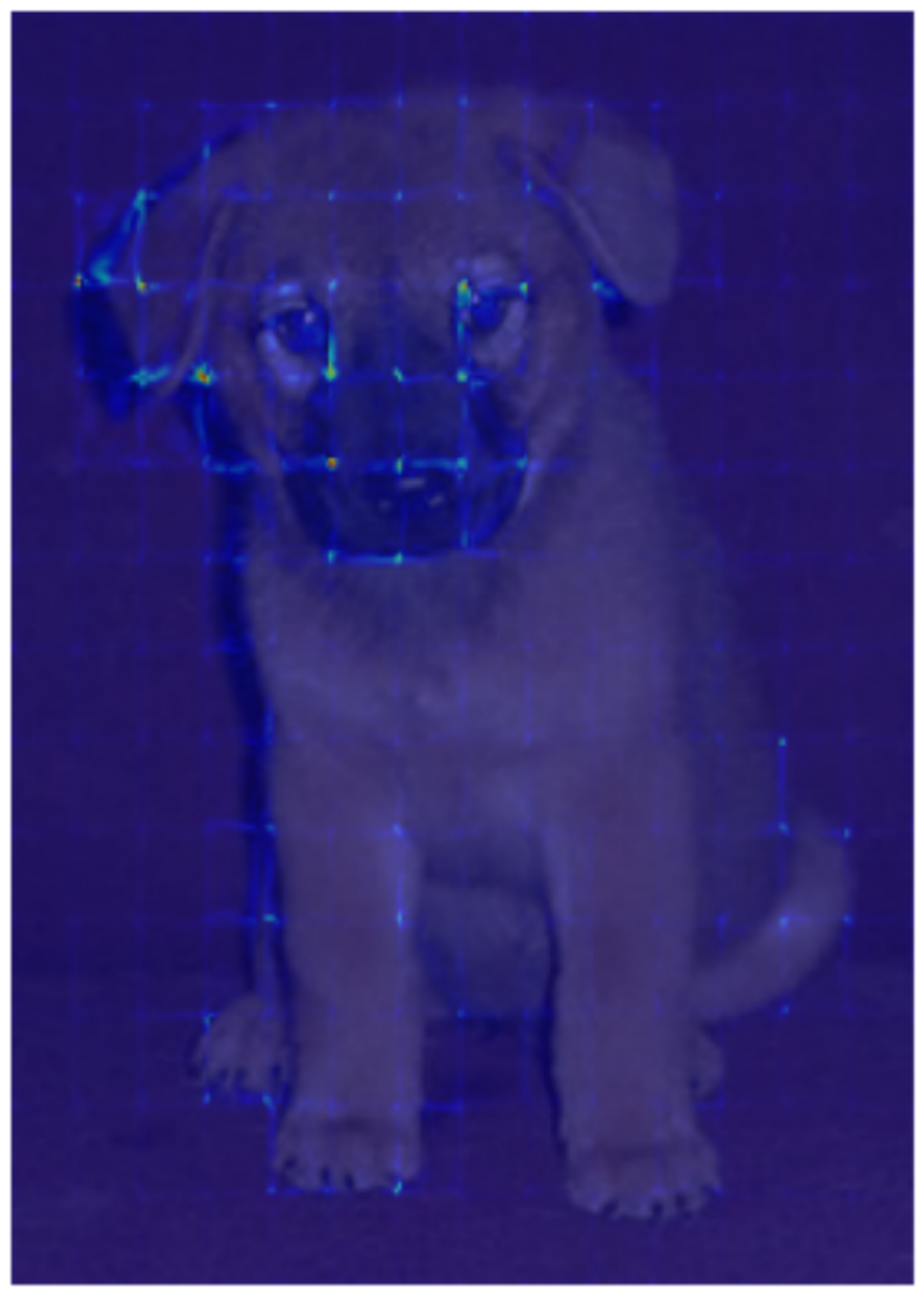} &
\includegraphics[width=0.13\textwidth]{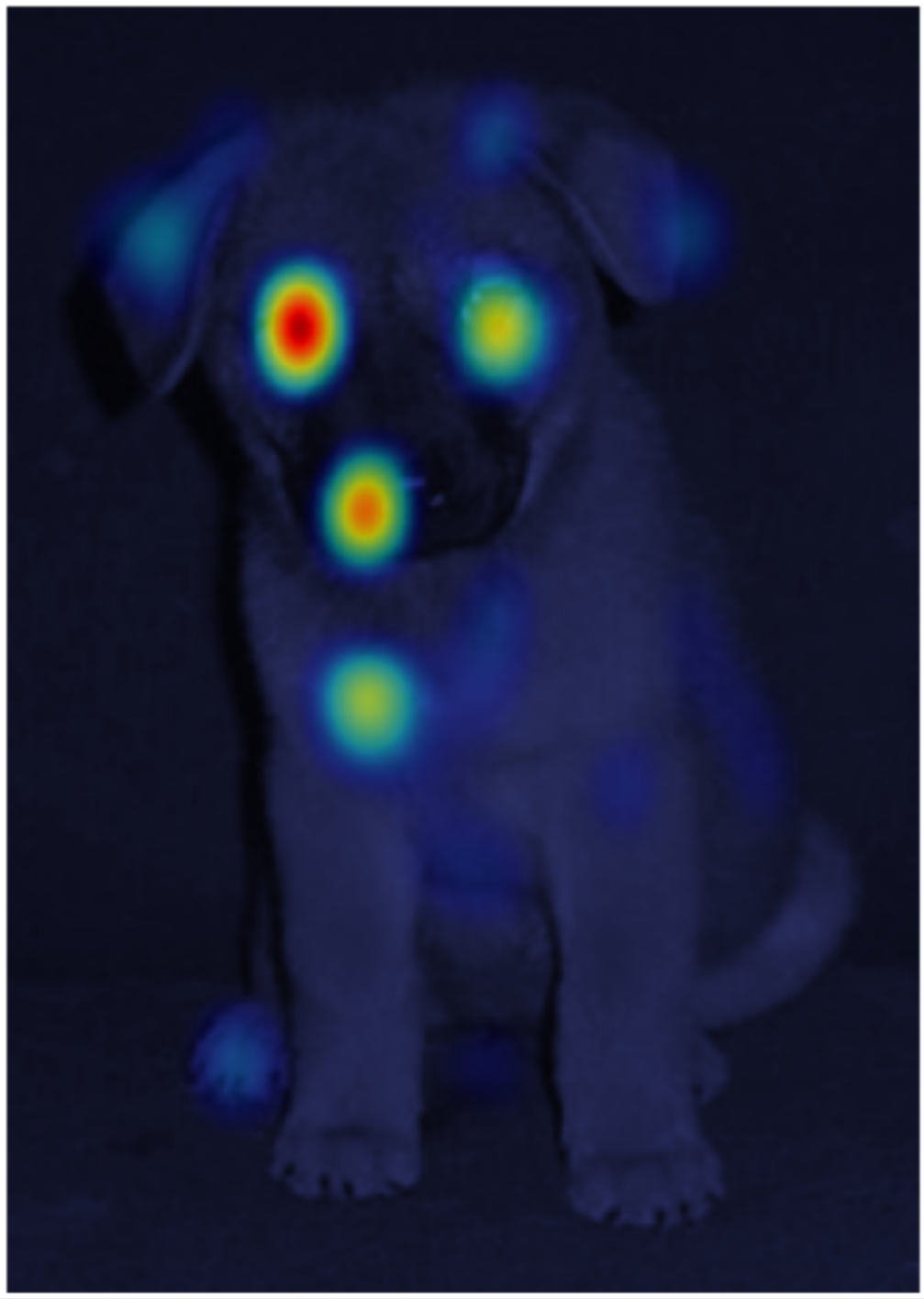} \\[2pt]

\end{tabular}

\caption{Qualitative comparison of explanations generated by baseline XAI methods and CA-LIG using a MAE model. Warmer colors denote regions with higher positive relevance, while cooler colors indicate lower relevance. Images are taken from the ASIRRA dataset \cite{elson2007asirra}.}

\label{fig:baseline_comparison1}
\end{figure*}

\begin{figure*}[t]
\centering
\setlength{\tabcolsep}{2pt}
\renewcommand{\arraystretch}{0}

\begin{tabular}{cccc}
\subcaptionbox{Original Input}[0.24\textwidth]{%
    \includegraphics[width=0.17\textwidth]{cat.jpg}
} &
\subcaptionbox{ CA-LIG Explanation}[0.2\textwidth]{%
    \includegraphics[width=0.17\textwidth]{cat_ours.png}
} &
\subcaptionbox{ Positive Attribution (helps prediction)}[0.24\textwidth]{%
    \includegraphics[width=0.17\textwidth]{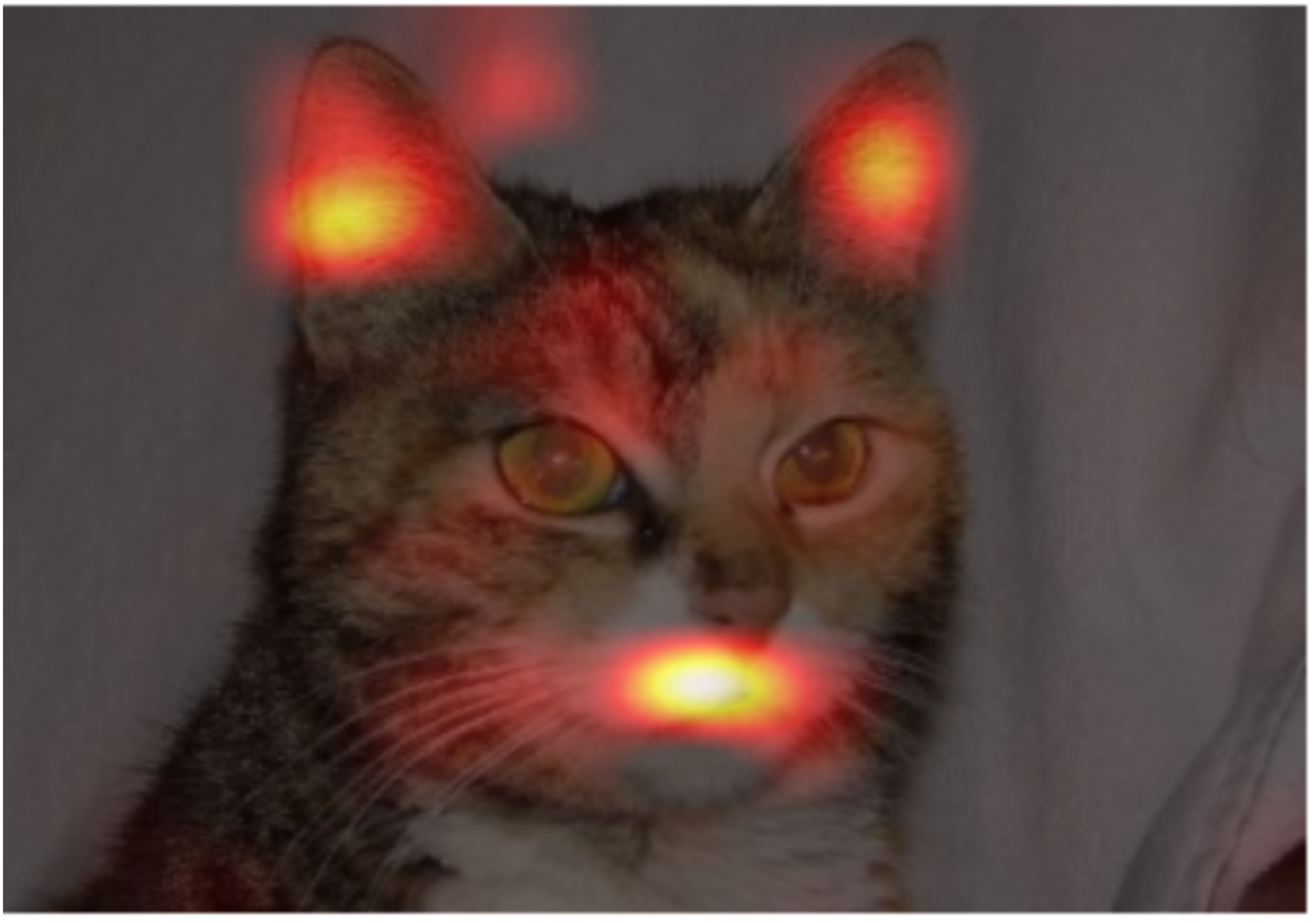}
} &
\subcaptionbox{ Negative Attribution (hinders prediction)}[0.24\textwidth]{%
    \includegraphics[width=0.17\textwidth]{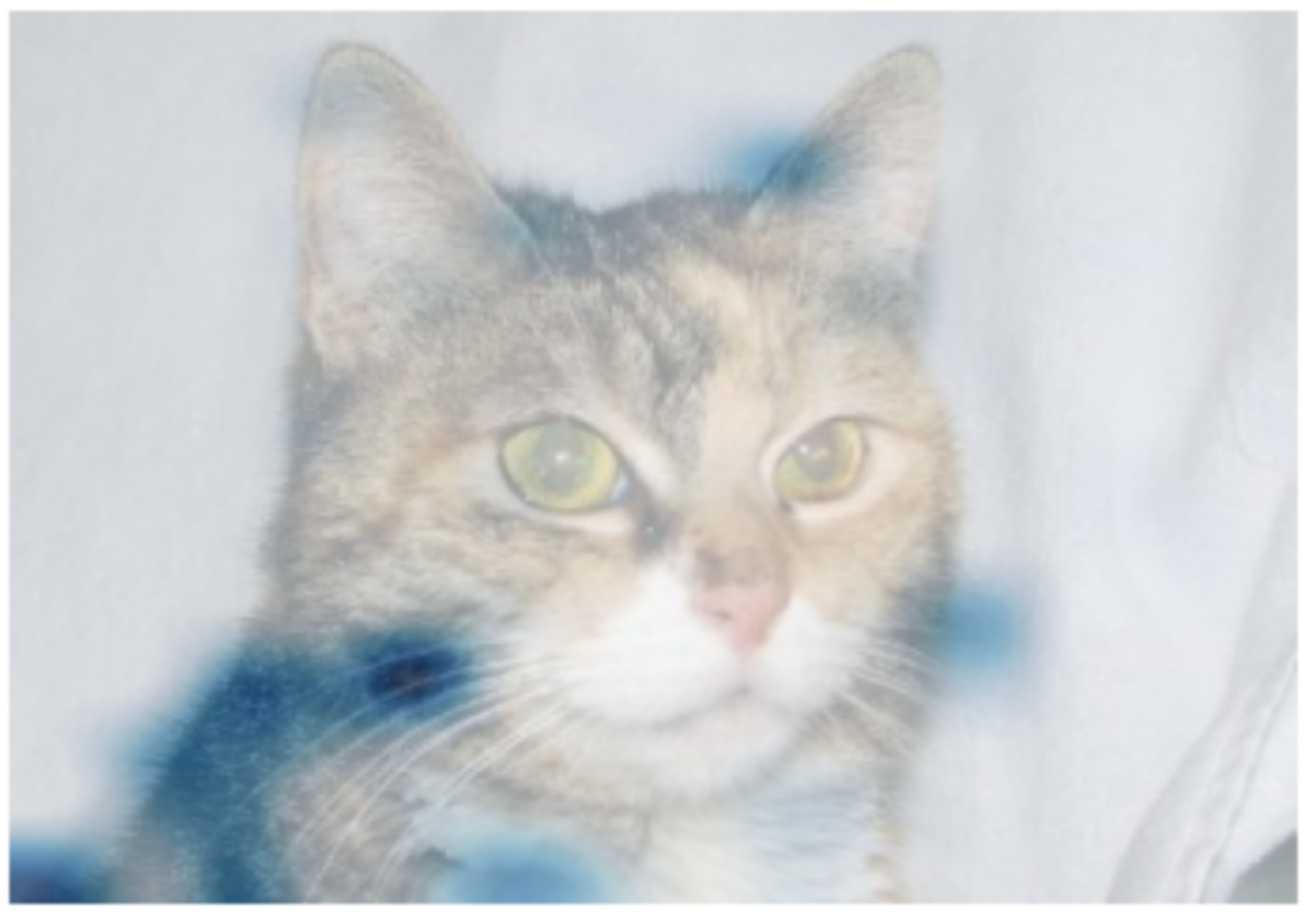}
} \\
\end{tabular}

\caption{Example of an explanation generated using CA-LIG for a prediction made by MAE model. (a) Original input image, (b) CA-LIG explanation heatmap, (c) positively attributed regions, and (d) negatively attributed regions. Warmer colors indicate stronger relevance.}
\label{fig:positive vs negative relevance}
\end{figure*}

\noindent
The CA-LIG attribution map highlights how BERT-Large captures semantically dependent tokens, as shown in Figure~\ref{Fig:Christian class}, even when these tokens appear far apart in the input sequence, demonstrating CA-LIG's ability to capture long-range contextual structure. Strong positive attributions on tokens such as \textit{God}, \textit{bible}, \textit{timing}, \textit{history}, \textit{world}, \textit{believers}, \textit{christ}, and \textit{close} indicate that CA-LIG identifies the theological and eschatological semantics that characterize the ``Christian'' class in the 20 Newsgroups dataset. Importantly, CA-LIG assigns high relevance not only to isolated words but also to \textit{concept pairs} that co-occur across multiple clauses---for example, \textit{evidence} $\rightarrow$ \textit{bible}, \textit{return} $\rightarrow$ \textit{christ}, \textit{day} $\rightarrow$ \textit{lord}, and \textit{end} $\rightarrow$ \textit{world}. These relationships show that CA-LIG does not rely solely on surface-level frequency but instead aggregates layer-wise signals that encode deep contextual interactions spanning entire sentences. For instance, the model links ``evidence in the bible" with ``timing of the history of the world", capturing a multi-clause reasoning chain that reflects Christian doctrinal themes, tokens that are not adjacent but are contextually unified. Similarly, negative relevance on words such as \textit{thief} and \textit{night} shows that CA-LIG appropriately down-weights metaphorical elements that do not directly characterize the Christian topic category. 

Figure ~\ref{Fig:atheist class} illustrates that CA-LIG assigns strong positive relevance (green) to multiple interacting topic-defining tokens such as “atheist,” “belief,” “system,” “lack belief,” and “gods,” which together establish the document’s ideological orientation. Instead of relying on a single lexical cue, CA-LIG captures how belief-related concepts and negation structures (e.g., “lack belief,” “never seen convincing evidence”) collectively drive the prediction, while supporting tokens receive weaker relevance and semantically peripheral tokens remain neutral.
Figure~\ref{Fig:Negative Sample2} presents CA-LIG attributions for a negative IMDB review using BERT-large, where strong negative relevance (red) is assigned to interacting sentiment-bearing expressions such as “worst,” “lame,” “poor acting,” and “things wrong.” The explanation shows that negative sentiment emerges from interactions across multiple evaluative tokens rather than a single word, with neutral or context-setting tokens remaining unhighlighted.
Figures~\ref{Fig:hate Sample} and~\ref{Fig:not hate Sample2} illustrate CA-LIG attributions for an Amharic hate speech sample using XLM-R and AfroLM models, respectively. CA-LIG assigns strong negative relevance to explicitly abusive tokens (Figure 8), particularly those corresponding to “these stupid” and “should be excluded,” which form the semantic core of the hate expression. Contextually reinforcing modifiers receive additional relevance, while syntactically necessary but semantically neutral tokens remain white.

These results demonstrate that CA-LIG explains model predictions through contextual interactions among semantically aligned tokens, capturing compositional meaning across clauses, sentiment expressions, and target–intent structures across models, domains, and languages. By combining layer-wise Integrated Gradients with attention-gradient fusion and context propagation, CA-LIG captures long-distance relevance patterns that reflect the hierarchical structure of the model’s reasoning, highlighting global semantic coherence rather than relying on local tokens alone.

In \emph{vision tasks}, input images are resized to $224 \times 224$ and split into distinct $16 \times 16$ patches, which are then linearly embedded into a sequence that includes a \texttt{[CLS]} token; a classifier head employs its terminal representation. Context is as crucial for vision tasks as it is for language, yet many XAI attribution methods overlook it, leading to fragmented or noisy explanations. Figure~\ref{fig:positive vs negative relevance} illustrates how the CA-LIG approach addresses this limitation by grounding explanations in semantically coherent features. Unlike baseline methods that highlight scattered pixel intensities, CA-LIG emphasizes relationally meaningful regions (e.g., the cat’s ears, eyes, and muzzle) that are directly related to object identity. Positive attributions capture features that drive the prediction (eyes, nose, whiskers), as shown in Figure~\ref{fig:positive vs negative relevance}c, while negative attributions highlight distracting regions (background textures, fur patterns), as shown in Figure~\ref{fig:positive vs negative relevance}d, offering counterfactual insights into model uncertainty. This contextual coherence improves interpretability by aligning with human reasoning, enhances robustness against noise and adversarial perturbations, and aids generalization by disambiguating class-relevant from irrelevant features. Overall, CA-LIG demonstrates that context is a universal requirement for explainability, bridging NLP and vision under a unified framework for context-aware XAI.

\subsection{Evaluation}
\subsubsection{Qualitative Evaluation}
\begin{figure}[ht]
\centering

\label{fig:xai_comparison}

\begin{minipage}{2cm}
\raggedright
XAI
\end{minipage}%
\begin{minipage}{5cm}
\raggedright
Explanation outputs
\end{minipage}\\[0.3em]

\begin{minipage}{1.9cm}
A-last \cite{hollenstein2021relative}
\end{minipage}%
\begin{minipage}{6.7cm}
\includegraphics[width=\textwidth]{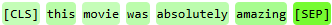}
\end{minipage}\\[0.3em]

\begin{minipage}{2.1cm}
A-Rollout \cite{abnar2020quantifying}
\end{minipage}%
\begin{minipage}{6.4cm}
\includegraphics[width=\textwidth]{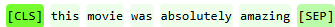}
\end{minipage}\\[0.3em]

\begin{minipage}{1.5cm}
IxG \cite{shrikumar2017learning}
\end{minipage}%
\begin{minipage}{7cm}
\includegraphics[width=\textwidth]{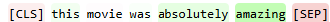}
\end{minipage}\\[0.3em]

\begin{minipage}{1.5cm}
LRP \cite{bach2015pixel}
\end{minipage}%
\begin{minipage}{7cm}
\includegraphics[width=\textwidth]{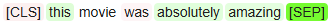}
\end{minipage}\\[0.3em]

\begin{minipage}{1.5cm}
IG \cite{sundararajan2017axiomatic}
\end{minipage}%
\begin{minipage}{7cm}
\includegraphics[width=\textwidth]{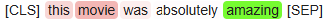}
\end{minipage}\\[0.3em]

\begin{minipage}{1.6cm}
\textbf{CA-IG}last (Ours)
\end{minipage}%
\begin{minipage}{7cm}
\includegraphics[width=\textwidth] {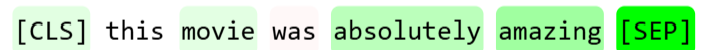}
\end{minipage}\\[0.3em]

\begin{minipage}{1.6cm}
\textbf{CA-LIG} (Ours)
\end{minipage}%
\begin{minipage}{7cm}
\includegraphics[width=\textwidth]{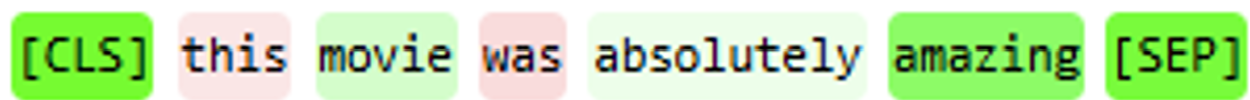}
\end{minipage}
\caption{Qualitative comparison of explanations produced by baseline XAI methods and CA-LIG using a BERT-base model. Brighter green indicates stronger positive relevance, red indicates negative relevance, and white represents neutral tokens.}
\label{Fig:baseline CA-LIG}
\end{figure}

Figure~\ref{Fig:baseline CA-LIG} presents a qualitative comparison of explanation visualizations for text classification. The baseline methods often emphasize irrelevant or dispersed tokens, resulting in noisy and less interpretable explanations. In contrast, our CA-LIG approach provides sharper and more focused attributions.

\subsubsection{Quantitative Evaluation}
We evaluate explanation quality on the \textit{Movie Reviews rationale benchmark} \cite{zaidan2007using, deyoung2019eraser}, which provides human-annotated rationale tokens for assessment. We report \textbf{token-F1}, measuring the overlap between the top-ranked tokens and the gold rationales \cite{chefer2021transformer}, using a length-normalized evaluation that selects the top-$p$\% of tokens ($p \in \{5,10,15,20,30,40,50\}$), with a minimum of five tokens per instance. Figure \ref{Fig:token-F1 evaluation} shows that while all methods improve as more tokens are included, our CA-LIG approach consistently achieves higher token-F1 than the baselines.

In the vision task, we assess the faithfulness of explanations using perturbation-based AUC, evaluated through the insertion and deletion of the most important patches, results presented in Table~\ref{Table: Perturbation AUC evaluation}. The image patches are ranked by importance and then progressively inserted from a blurred baseline or deleted from the original input image. The perturbation curves are computed using a length-normalized procedure that updates 5\% of the patches per step, over 20 steps ranging 0–100\%. A faithful explanation produces a rapid increase in confidence in insertion (larger AUC) and a decrease in confidence in deletion (smaller AUC). In the cat and dog examples, our method consistently emphasizes class-defining regions (e.g., eyes, muzzle), resulting in sharper insertion gains and stronger deletion declines compared to baselines.

\begin{table*}[t]
\centering
\setlength{\tabcolsep}{6pt}
\begin{tabular}{llcccccc}
\toprule
\multicolumn{2}{c}{Perturbation} & Class & GradCAM \cite{selvaraju2017grad} & LRP-$\epsilon$ \cite{bach2015pixel} & LRP (Attn+Grad) \cite{chefer2021transformer} & IG \cite{sundararajan2017axiomatic} & Ours \\
\midrule
\multirow{1}{*}{ Insertion $\uparrow$} 
&  & Predicted & 0.424 & 0.448 & 0.567 & 0.473 & \textbf{0.617} \\
\multirow{1}{*}{ Deletion $\downarrow$} 
&  & Predicted & 0.374 & 0.437 & 0.237 & 0.346 & \textbf{0.215} \\
\bottomrule
\end{tabular}
\caption{Perturbation-based AUC evaluation of explanation faithfulness. Higher insertion AUC (↑) indicates faster confidence recovery when important patches are added to a blurred baseline, while lower deletion AUC (↓) reflects faster confidence drop when critical patches are removed from the original image.}
\label{Table: Perturbation AUC evaluation}
\end{table*}

\subsection{Case Study: Layer-wise Sensitivity Analysis of CA-LIG}
Transformer-based architectures, such as BERT, demonstrate a hierarchical progression of linguistic representations across their layers. Empirical analyses have shown that the early layers (1--4) primarily encode surface-level syntactic features and grammatical relations \citep{tenney2019bert,hewitt2019structural,goldberg2019assessing,aoyama2022probe,ferrando2022measure}, the middle layers (5--8) capture semantic meaning and contextual dependencies \citep{liu2019linguistic,clark2019bert,nauta2023anecdotal,liu2024cunliang}, and the deeper layers (9--12) consolidate sentence-level information to support task-specific decision making, particularly through refinement of the \texttt{[CLS]} representation \citep{sun2019fine,ethayarajh2019contextual,kovaleva2019revealing}.

To empirically investigate how our proposed CA-LIG framework aligns with this hierarchical organization, we conducted a case study on the input sentence “This movie was absolutely amazing” using a BERT-base model fine-tuned on the IMDB sentiment classification dataset. The goal of this study is to trace how token-level relevance evolves across layers and how it interacts with classifier alignment and attention dynamics.

For each Transformer layer \(L\), we extracted three complementary interpretability signals: (1) CA-LIG token-level relevance scores; (2) classifier contribution computed by projecting the hidden state \(h_L\) of each token onto the final classification layer as \(C_L = h_L \cdot W_c + b_c\); and (3) mean attention to the \texttt{[CLS]} token, averaged across all heads, to assess shifts in the focus of aggregation.

As shown in Figure~\ref{Fig:Case Sensitivty Analysis}, the layer-wise sensitivity patterns exhibit three distinct phases, aligned with the structural-functional hierarchy of BERT described in prior studies.

In the \textit{surface feature extraction stage} (Layers 1--4), CA-LIG relevance scores remain low, reflecting minimal token-level differentiation. A modest increase is observed at Layer~4, coinciding with an increase in the classifier contribution, suggesting the early stages of representational alignment with the decision space \citep{rogers2020}. Attention to the \texttt{[CLS]} token remains relatively stable, indicating that attribution changes are driven primarily by representational transformation rather than attention reweighting. Attributions in this stage are uniformly distributed across tokens, with a limited emphasis on either syntactic or sentiment-bearing terms, consistent with the role of these layers in capturing POS and dependency structures \citep{hewitt2019structural,goldberg2019assessing}.

In the \textit{contextual reasoning stage} (Layers 5--8), CA-LIG scores increase sharply, peaking at Layer~8, while classifier contributions remain elevated. Token-level relevance patterns reveal a de-emphasis on functional terms such as ``was'' and increased attribution toward sentiment-rich tokens like ``absolutely'' and ``amazing'', as shown in Figure \ref{Fig:Layer-wise explanation}. This observation aligns with findings that these middle layers encode compositional semantics and contextual refinements \citep{liu2019linguistic,clark2019bert, peters2018}. Despite this semantic refinement, attention to \texttt{[CLS]} remains a kind of stable, further supporting that internal representational adjustments—not shifts in attention flow—account for the observed relevance dynamics.

In the \textit{decision consolidation} stage (Layers 9--12), CA-LIG scores briefly dip at Layer~10 before surging to their highest level at Layer~12. This peak co-occurs with the maximum classifier contribution, indicating the final integration of task-relevant evidence into the \texttt{[CLS]} embedding, a pattern in line with deep-layer specialization for decision-making \citep{sun2019fine,kovaleva2019revealing}. Here, relevance becomes sharply concentrated on the sentiment-bearing tokens, highlighting their role in determining model predictions. Despite this semantic convergence, attention to \texttt{[CLS]} remains relatively stable, suggesting that the model relies more on internal representational refinement than attention flow adjustments.

Overall, this analysis reveals that CA-LIG effectively captures the hierarchical progression of interpretability across Transformer layers. The observed transition points at Layers~4 and~8 align with prior probing studies~\citep{tenney2019bert,liu2019linguistic}, supporting the interpretive utility of CA-LIG in determining how contextual and task-relevant signals emerge within the model. Notably, CA-LIG token relevance trends closely mirror classifier contributions at key semantic layers, confirming its ability to track decision-aligned transformations. The weak alignment with [CLS] attention highlights that interpretability emerges from representational shifts rather than attention redistribution alone. These insights reinforce the view that interpretability techniques should account for the dynamic flow of information across layers, rather than relying solely on final-layer explanations.

\section{Limitations and Future Work} \label{sec:limitations}
While the CA-LIG framework provides a comprehensive and context-aware approach to interpreting Transformer decoder-based models, some limitations remain. First, the framework is designed for encoder-only models and has not yet been experimented with decoder-based language models. Second, the fusion coefficient $\lambda$ is manually tuned, requiring adaptive or learnable strategies. Third, while CA-LIG demonstrates strong performance on text-based tasks, its evaluation in the vision domain remains limited. More extensive experiments involving multiple vision models, tasks, and diverse datasets are required to fully assess its effectiveness in visual settings. Additionally, this study does not address multimodal Transformer explainability, in which cross-modal attention pathways introduce distinct interpretability challenges. In future work, we plan to address these limitations by extending CA-LIG to vision-based, decoder-based, and multimodal Transformer models, developing adaptive fusion mechanisms, and conducting broader evaluations across language, vision, and multimodal domains.

\section{Conclusion} \label{sec:conclusion}
Despite their strong performance, Transformer models remain difficult to interpret. Most existing XAI methods often rely on final-layer attributions, focus on either local gradients or global attention patterns, lack explicit context-awareness, and fail to preserve relevance consistently across layers. These limitations prevent faithful and structurally coherent explanations of Transformer decision-making.

In this study, we introduced the \textbf{CA-LIG Framework}, a unified and hierarchical attribution framework designed to address these gaps. CA-LIG differs from conventional single-layer attribution by applying \emph{Layer-wise Integrated Gradients} at every Transformer block, thereby revealing how token relevance emerges, evolves, and stabilizes as the network deepens. By incorporating \emph{class-specific attention gradients}, CA-LIG captures the contextual dependencies and structural information flow that shape model predictions. The framework fuses these complementary signals through a context-aware integration mechanism and aggregates them using a relevance rollout procedure that preserves attribution stability and conserves the relevance across layers. Extensive experiments on sentiment analysis, low-resource hate speech detection, long-form and multiclass document classification, and image classification tasks demonstrate that CA-LIG consistently outperforms established XAI techniques. Both quantitative evaluations and qualitative analyses confirm that CA-LIG produces explanations that are not only more reliable but also more aligned with the hierarchical computation that defines Transformer architectures. 

CA-LIG enhances Transformer explainability by integrating completeness, context-awareness, and hierarchical fidelity. By tracing relevance across layers and unifying token-level and structural dependencies, it generates explanations that are more faithful, coherent, and aligned with human reasoning. This study presents a meaningful step toward building transparent, interpretable, and trustworthy Transformer-based models.

\appendix
\section{Methodological Comparison of CA-LIG with Prior Methods} \label{appendix A}

Methodologically, the CA-LIG Framework occupies a distinct space within the spectrum of Transformer interpretability approaches. 
Table~\ref{tab:methodlogicalComparison}, in \ref{appendix A}, presents a comparison of CA-LIG with the most widely used Transformer interpretability methods.
\onecolumn

\begin{table*}[ht]
\centering
\small
\renewcommand{\arraystretch}{1.25}
\setlength{\tabcolsep}{4pt}

\begin{tabular}{p{2.5cm} p{3.2cm} p{3.1cm} p{3.3cm} p{4.4cm}}
\hline
\textbf{Aspect} &
\textbf{IG ~\cite{sundararajan2017axiomatic}} &
\textbf{Integrated Hessians  ~\cite{janizek2021explaining}} &
\textbf{Transformer-LRP ~\cite{chefer2021transformer}} &
\textbf{CA-LIG Framework (Ours)} \\
\hline

\textbf{Core Principle} &
Path-integrated gradients from baseline to input &
Second-order IG using Hessians for interaction effects &
LRP/Deep Taylor rules  &
Layer-wise IG fused with class-specific attention gradients \\

\textbf{Primary Objective} &
Token-level importance at input layer. &
Quantify pairwise feature interactions. &
Propagate and conserve relevance across attention and residuals. &
Produce hierarchical, context-aware token relevance across layers. \\

\textbf{Derivative Order} &
First-order &
Second-order &
Rule-based &
First-order IG with gradient-weighted attention integration \\

\textbf{Granularity} &
Input-level token salience &
Interaction matrix between feature pairs &
Single global input-level relevance map &
Per-layer token attribution + fused contextual map + multi-layer rollout \\

\textbf{Use of Attention} &
None &
None &
Implicit via LRP propagation rules &
Explicit class-specific attention \emph{gradients}. \\

\textbf{Layer-wise Tracking} &
None (only final output layer) &
None (interaction-focused) &
Relevance flows through layers &
Explicit LIG at each layer to model relevance trajectories. \\

\textbf{Relevance Conservation} &
Completeness guaranteed for IG &
No global conservation guarantees &
Strict conservation via LRP rules &
Approximate conservation across layers with normalized fusion \\

\textbf{Context Awareness} &
Weak (only at final layer) &
Indirect via interaction structure &
Partially context-aware via attention propagation rules &
Explicit capturing of structural context via attention gradients + LIG \\

\textbf{Output Representation} &
Single saliency map over tokens &
Interaction saliency matrix &
Single relevance heatmap &
Signed multi-layer relevance maps + fused final explanation \\

\textbf{Computational Cost} &
Low (only final layer) &
High (mixed Hessian integrals) &
Moderate &
Moderate–High (per-layer IG + attention gradients) \\

\textbf{Best Use Cases} &
General-purpose saliency &
Interaction analysis for important pairs &
Attention-heavy Transformer interpretability &
Hierarchical reasoning, context flow, fine-grained token relevance \\
\hline
\end{tabular}

\caption{Methodological comparison of IG, Integrated Hessians, Transformer-LRP, and the proposed CA-LIG framework, highlighting how CA-LIG combines layer-wise Integrated Gradients with attention-gradient fusion to produce hierarchical, context-aware explanations.}
\label{tab:methodlogicalComparison}
\end{table*}

\section{Supplementary Experimental Results} \label{appendix B}

\begin{figure*}[ht]
\centering
\includegraphics[width=\textwidth]{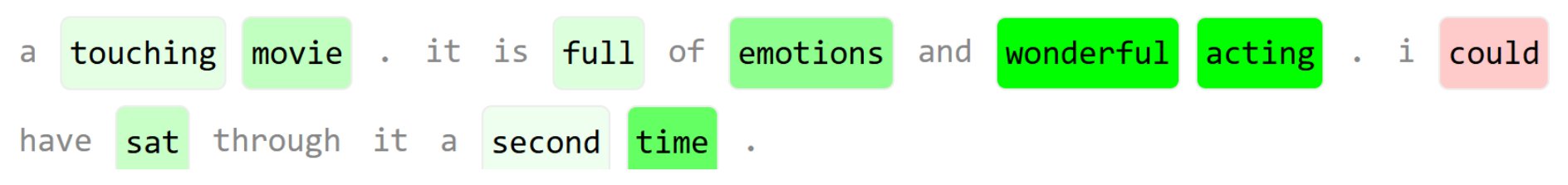}
\caption{CA-LIG token-level attributions for a positive IMDB review using BERT-Large. Brighter green indicates stronger positive evidence, red indicates negative relevance, and white denotes neutral tokens.}
\label{Fig:Positive Sample}
\end{figure*}

\begin{figure*}[ht]
\centering
\includegraphics[width=\textwidth]{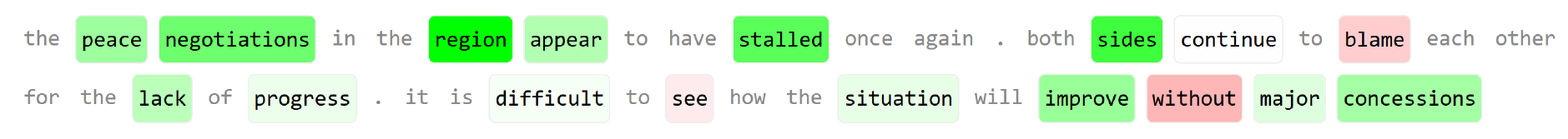}
\caption{CA-LIG token-level attributions for a \textit{document labeled politics class} from the 20 Newsgroups dataset using BERT-base. Brighter green tokens provide stronger positive evidence, lighter green indicates weaker support, red shows negative influence, and white denotes neutral relevance.}
\label{Fig: Positive Sample}
\end{figure*}

\begin{figure*}[ht]
\centering
\includegraphics[width=0.8\textwidth]{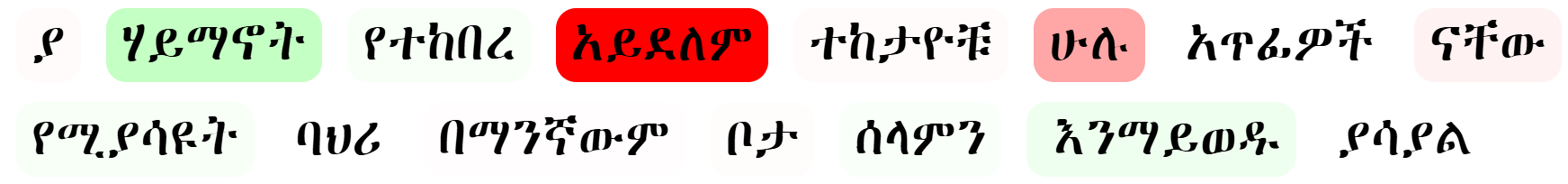}
\caption{Token-level attribution visualization generated by the CA-LIG using the AfroLM model on an Amharic hate speech sample. Brighter
Red tokens provide stronger negative evidence (hate); lighter red indicates weaker support; green shows positive influence; and white denotes neutral relevance. English translation: \textit{That religion is not respected, all its followers are destructive, and their behavior shows that they do not like peace anywhere.}}
\label{Fig:NonHate Sample}
\end{figure*}

\begin{figure*}[ht]
\centering
\includegraphics[width=\textwidth]{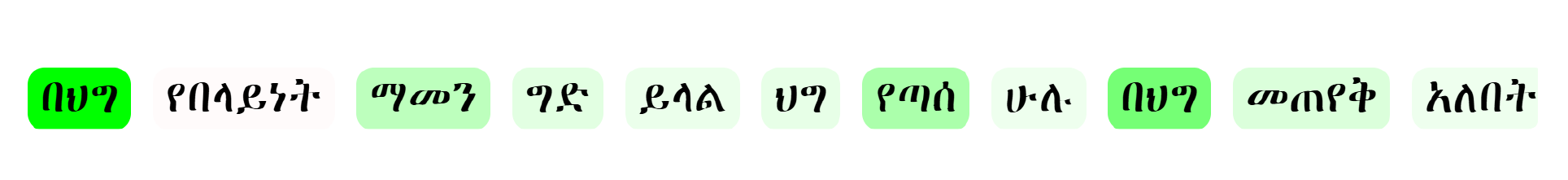}
\caption{Token-level attribution visualization generated by the CA-LIG using the XLM-R model on an Amharic hate speech sample. Brighter
green tokens provide stronger positive evidence (not hate), lighter green indicates weaker support, red shows negative influence, and white denotes neutral relevance. English translation: \textit{It is important to believe in the rule of law. Anyone who breaks the law must be held accountable.}}
\label{Fig:NonHate Sample}
\end{figure*}

\begin{figure*}[t]
\centering
\setlength{\tabcolsep}{2pt}
\renewcommand{\arraystretch}{0}

\begin{tabular}{cccc}
\subcaptionbox{Original Input}[0.24\textwidth]{%
    \includegraphics[width=0.17\textwidth]{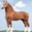}
} &
\subcaptionbox{ CA-LIG Explanation}[0.2\textwidth]{%
    \includegraphics[width=0.17\textwidth]{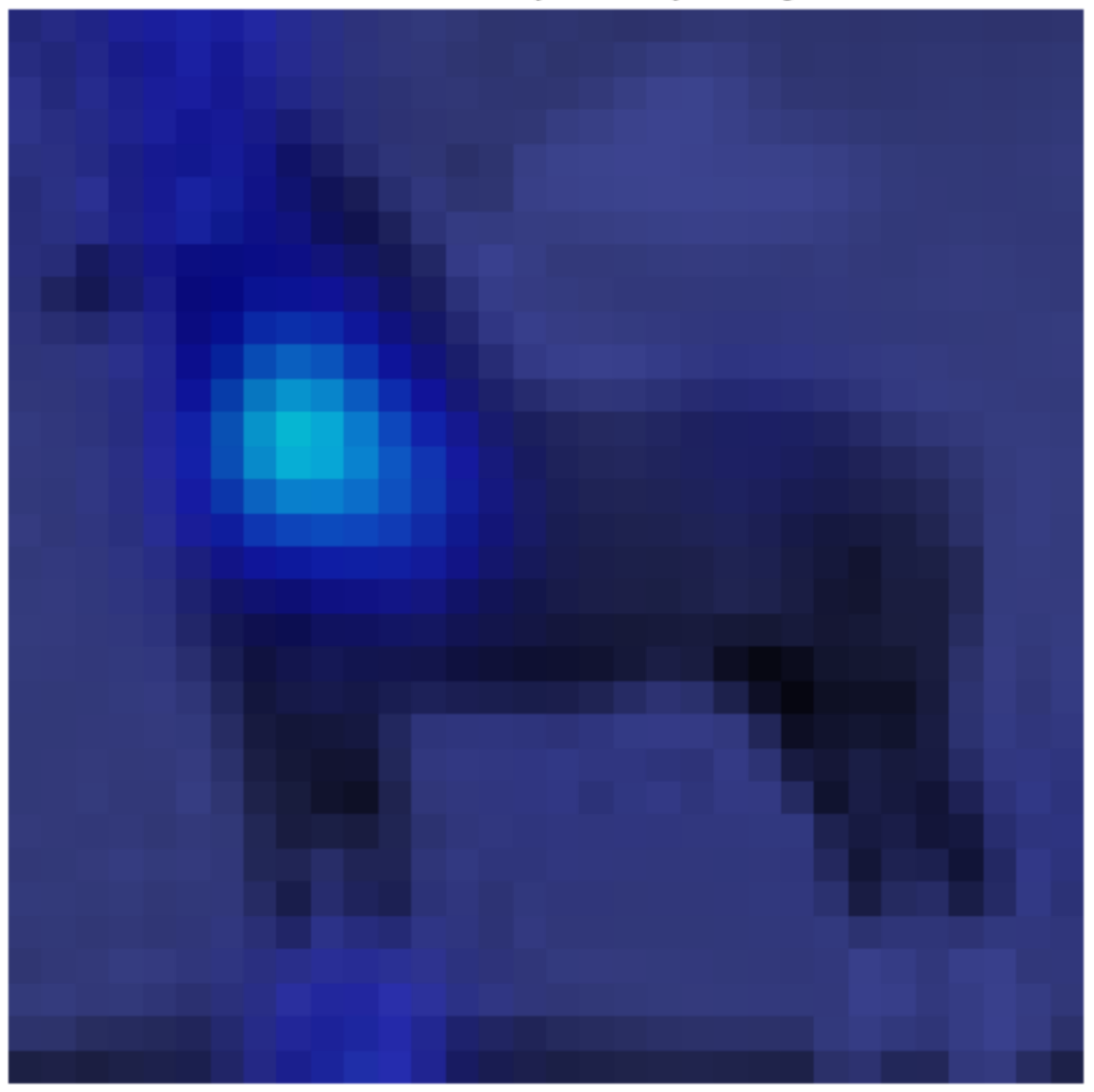}
} &
\subcaptionbox{ Positive Attribution (helps prediction)}[0.24\textwidth]{%
    \includegraphics[width=0.17\textwidth]{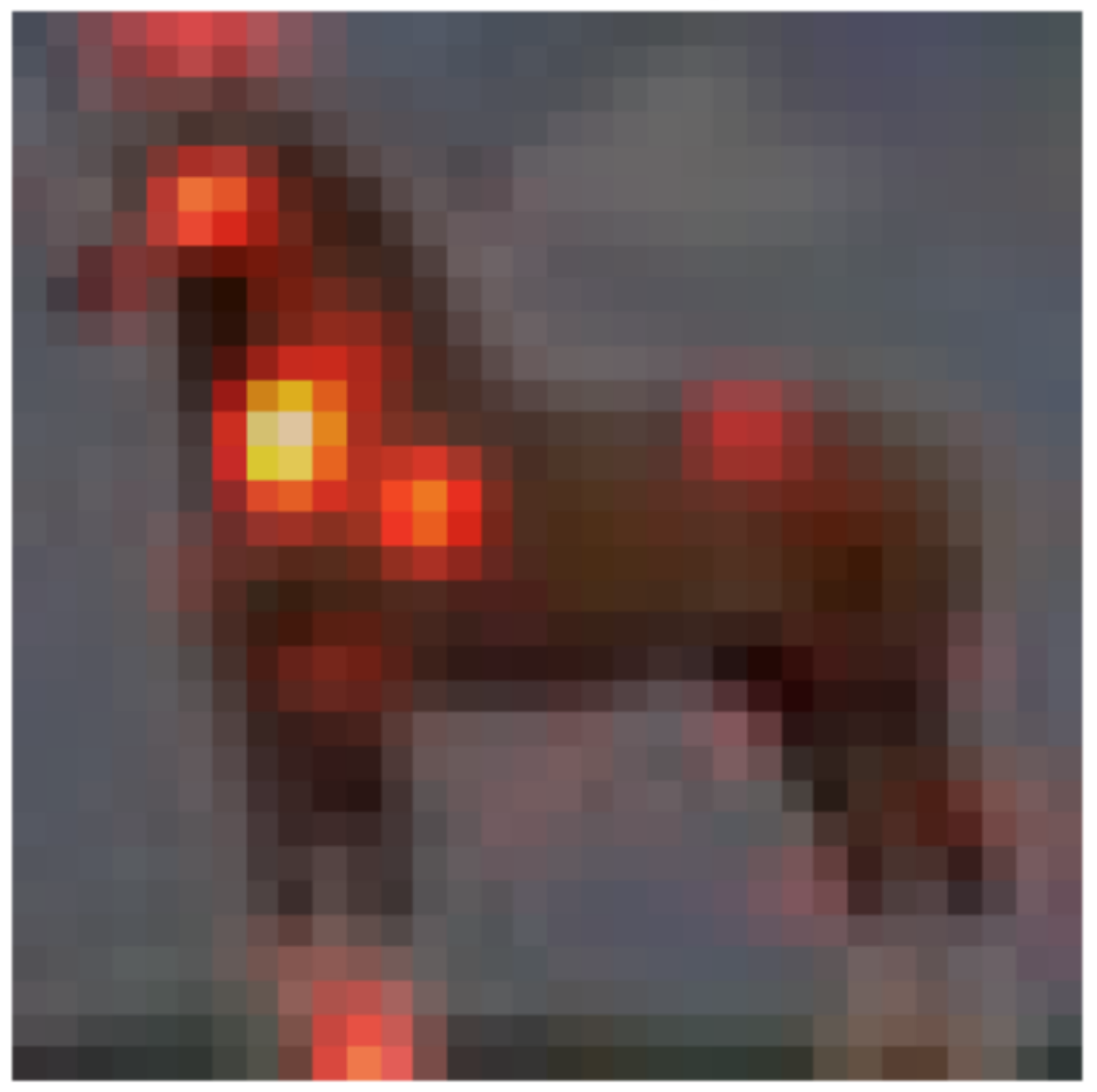}
} &
\subcaptionbox{ Negative Attribution (hinders prediction)}[0.24\textwidth]{%
    \includegraphics[width=0.17\textwidth]{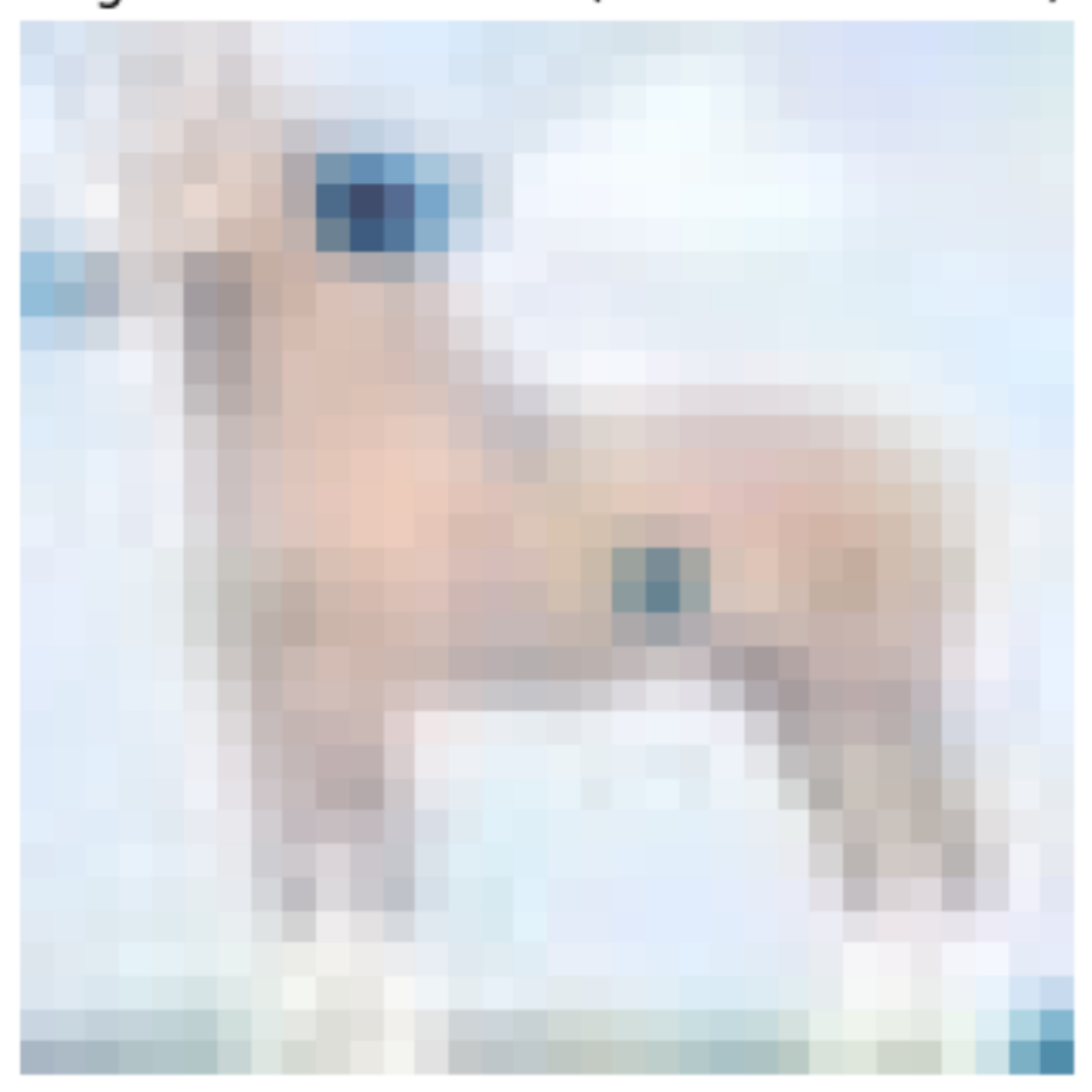}
} \\
\end{tabular}

\caption{Example of a CA-LIG explanation for a prediction made by the MAE model on the CIFAR-10 dataset. (a) Original input image, (b) CA-LIG explanation heatmap, (c) positively attributed regions, and (d) negatively attributed regions. Warmer colors indicate stronger relevance to the model’s prediction. The CA-LIG explanation indicates that the MAE model’s predictions are driven by contextually related and semantically meaningful object parts. As illustrated in (b) and (c), strong positive relevance is consistently assigned to the head, torso, and lower body/feet. In contrast, (d) indicates negative relevance to background and peripheral regions that do not contribute to the prediction. }
\label{fig:positive vs negative relevance}
\end{figure*}

\begin{figure*}[t]
\centering
\setlength{\tabcolsep}{2pt}
\renewcommand{\arraystretch}{0}

\begin{tabular}{cccccc}
\small\textbf{Input} & \small\textbf{Grad-CAM \cite{selvaraju2017grad}} & \small\textbf{LRP-$\epsilon$ \cite{bach2015pixel}} & \small\textbf{LRP(Attn+Grad) \cite{chefer2021transformer}} & \small\textbf{IG \cite{sundararajan2017axiomatic}} & \small\textbf{CA-LIG (Ours)} \\[2pt]

\includegraphics[width=0.16\textwidth]{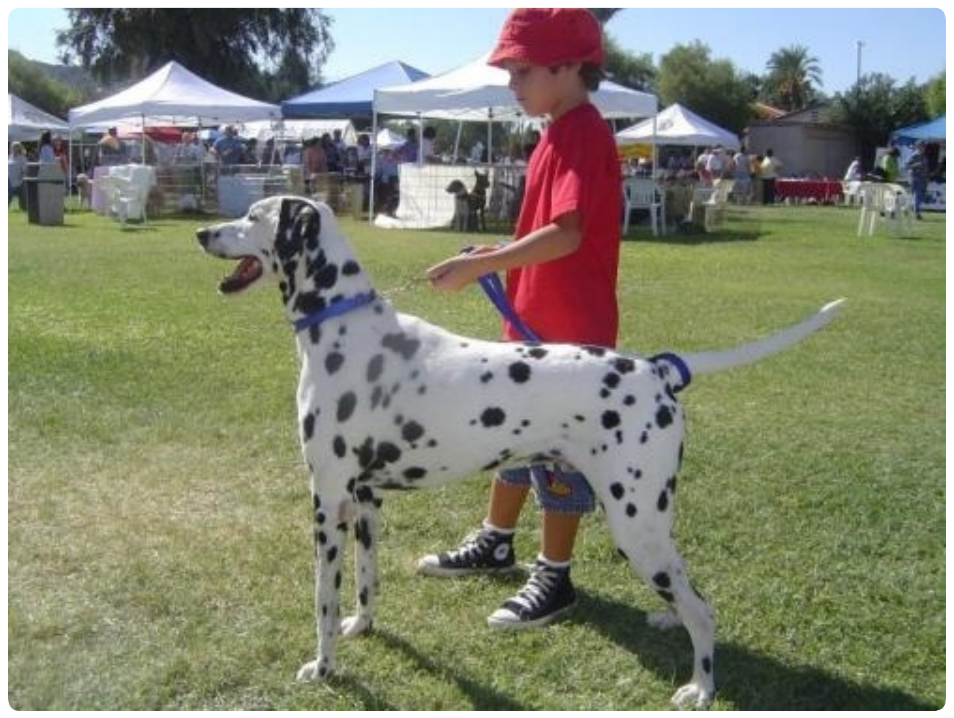} &
\includegraphics[width=0.16\textwidth]{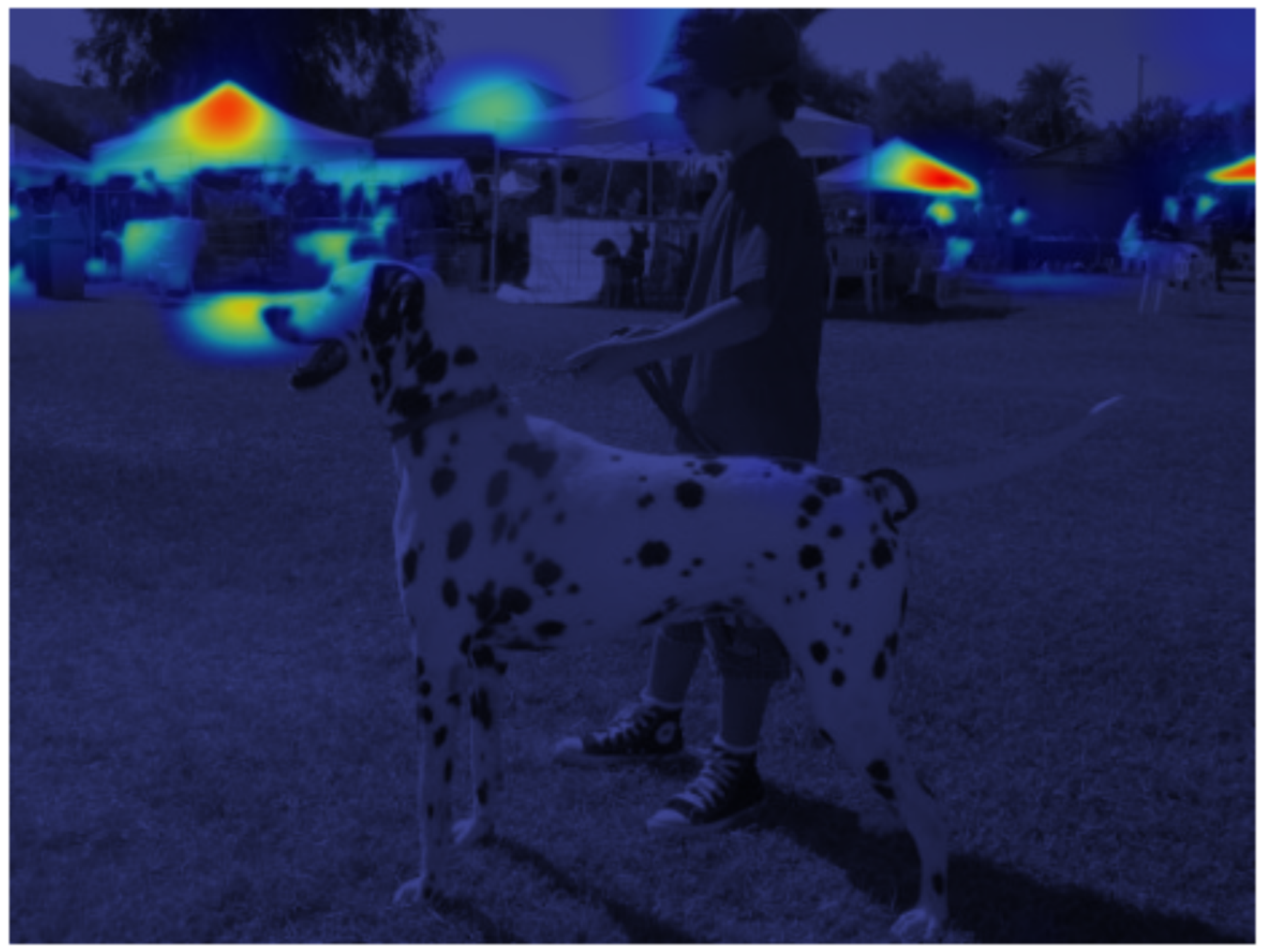} &
\includegraphics[width=0.16\textwidth]{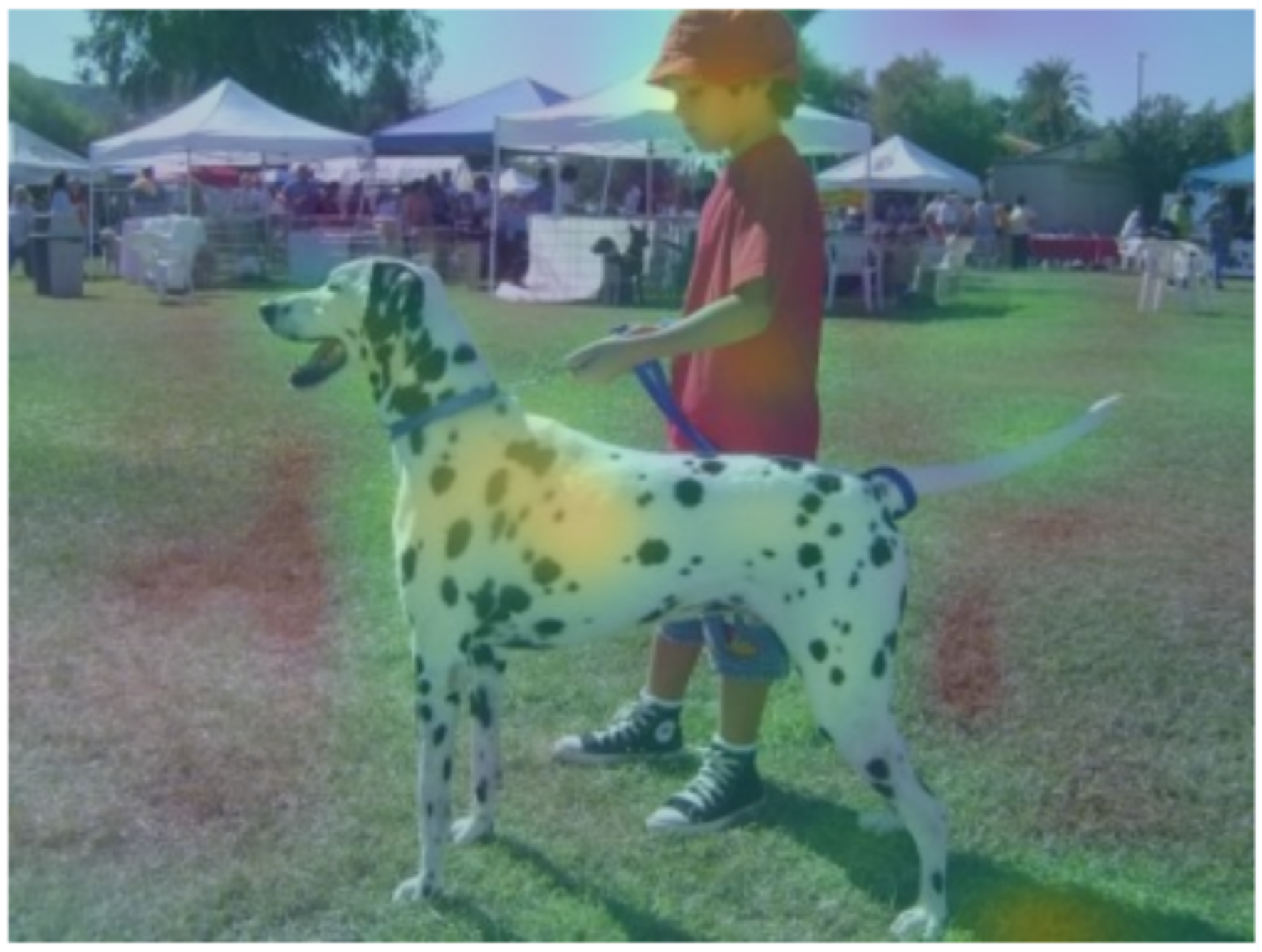} &
\includegraphics[width=0.16\textwidth]{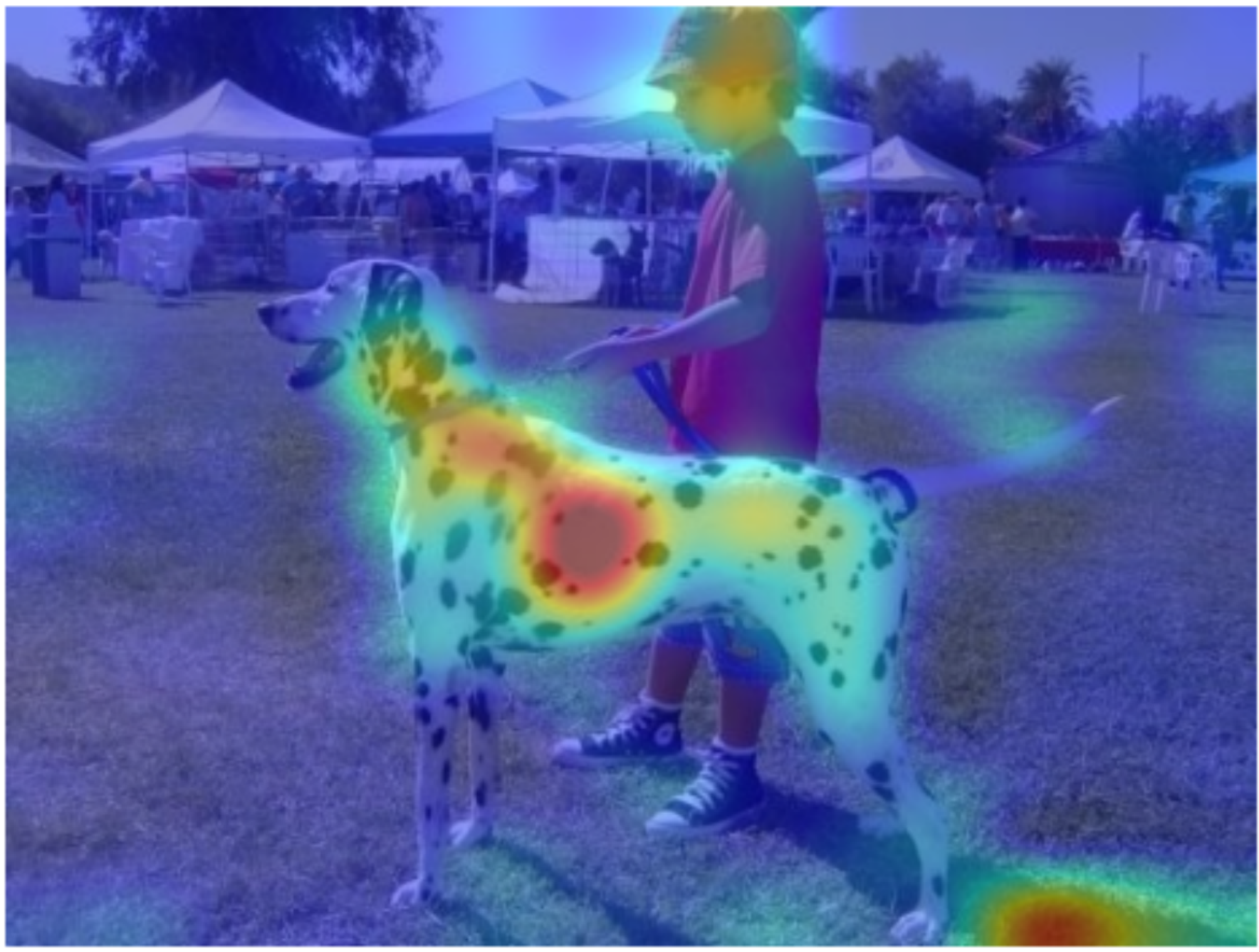} &
\includegraphics[width=0.16\textwidth]{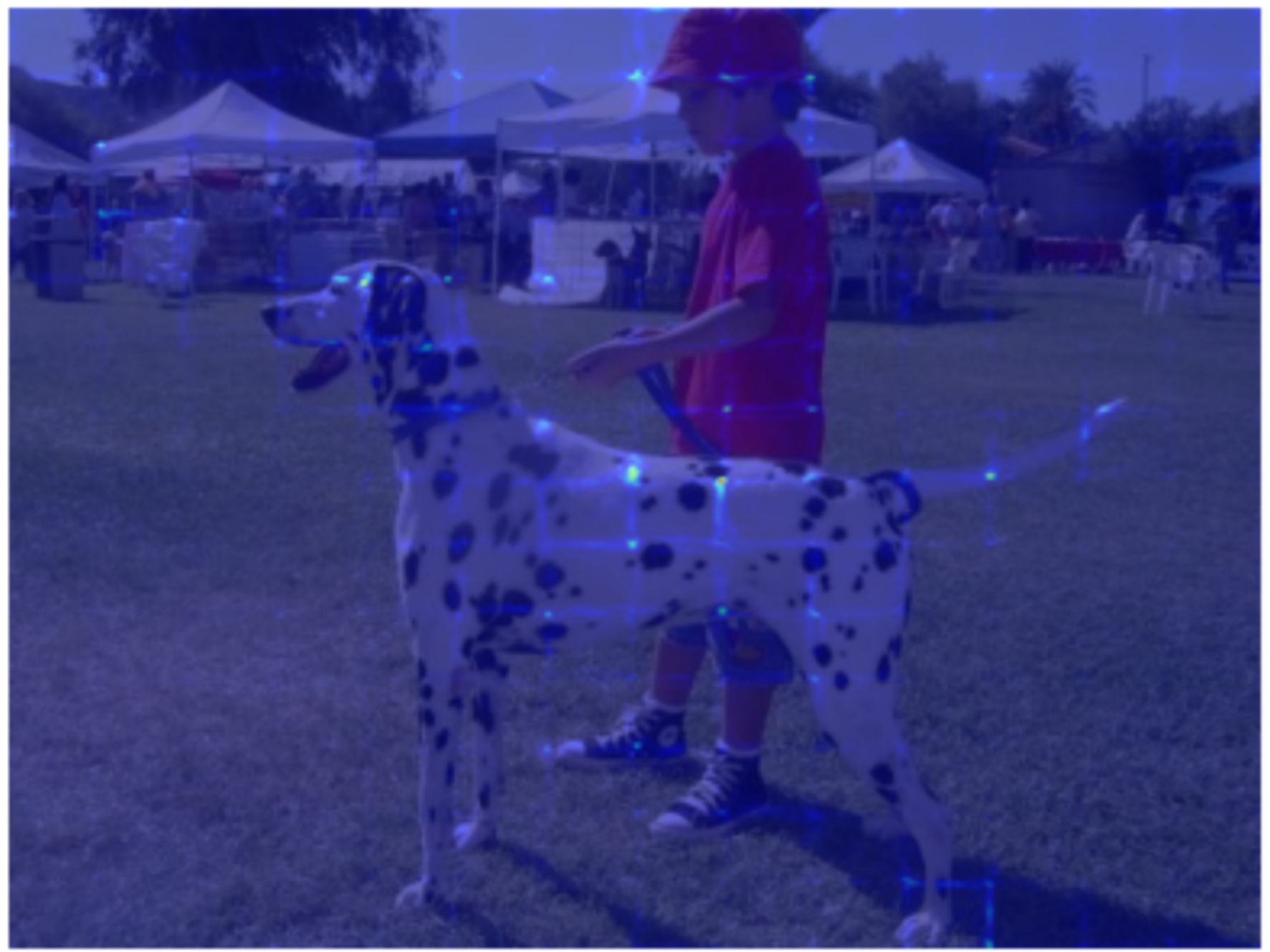} &
\includegraphics[width=0.16\textwidth]{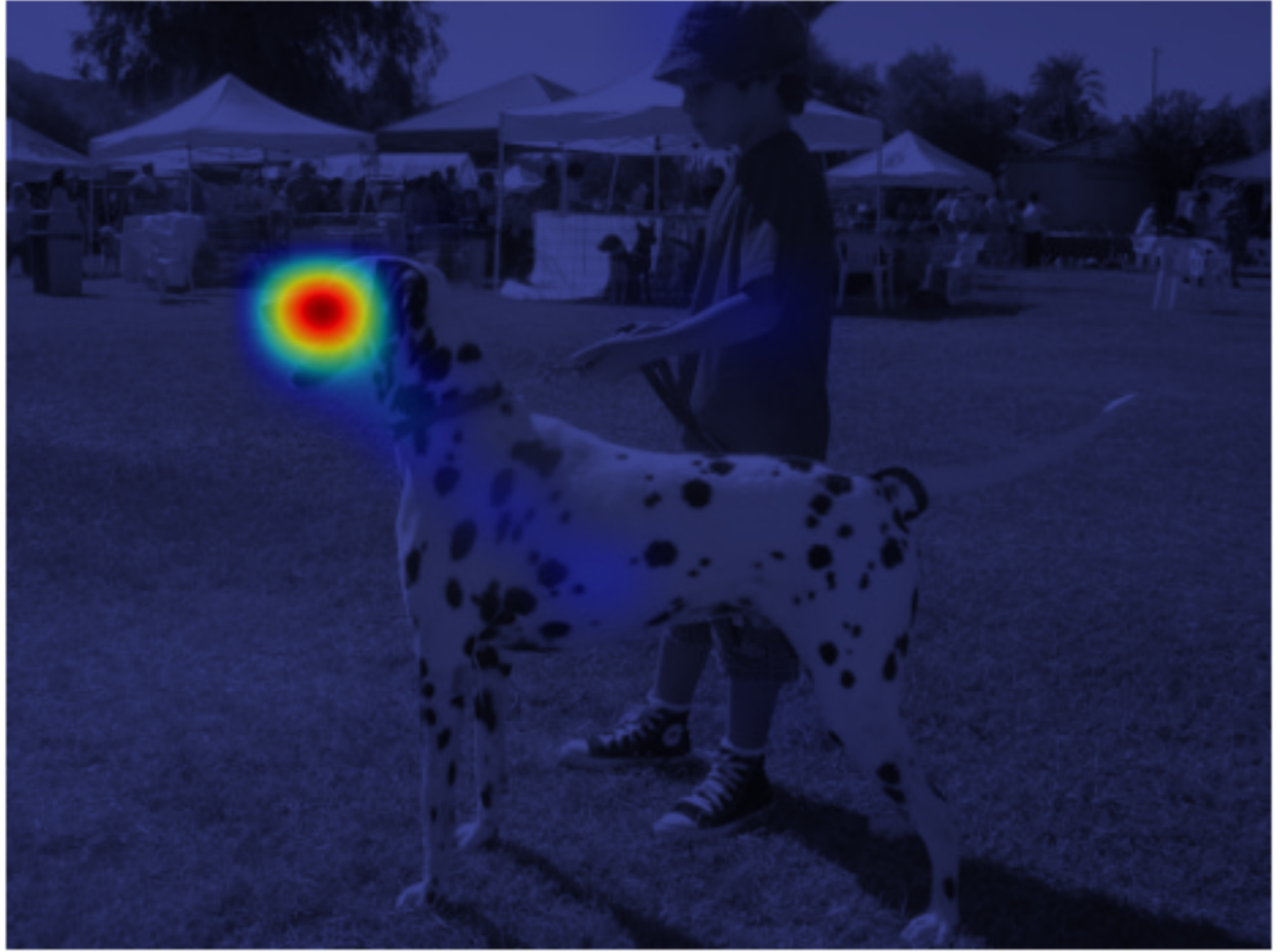} \\[2pt]

\includegraphics[width=0.16\textwidth]{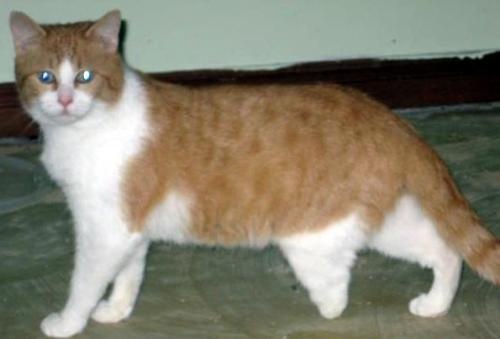} &
\includegraphics[width=0.16\textwidth]{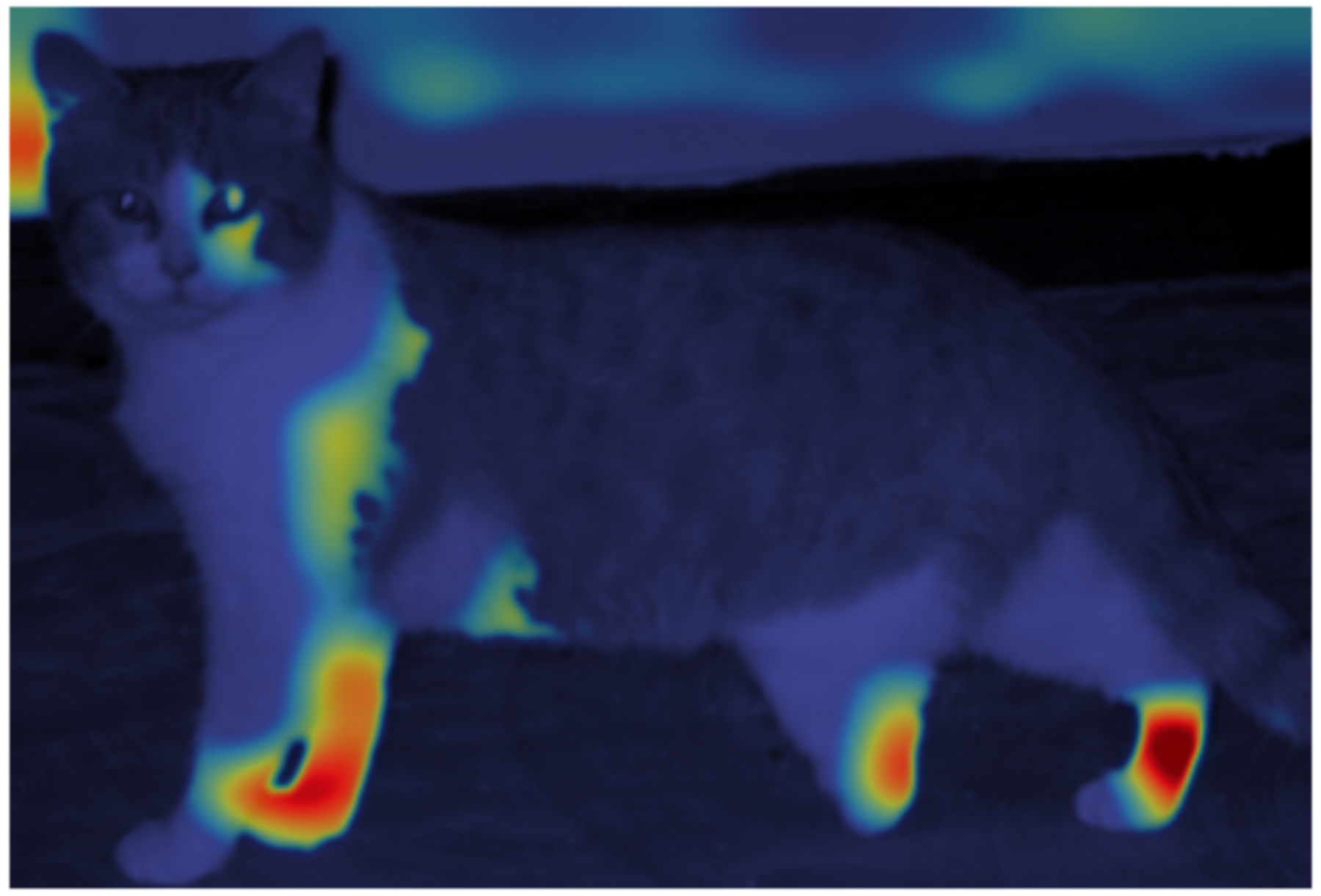} &
\includegraphics[width=0.16\textwidth]{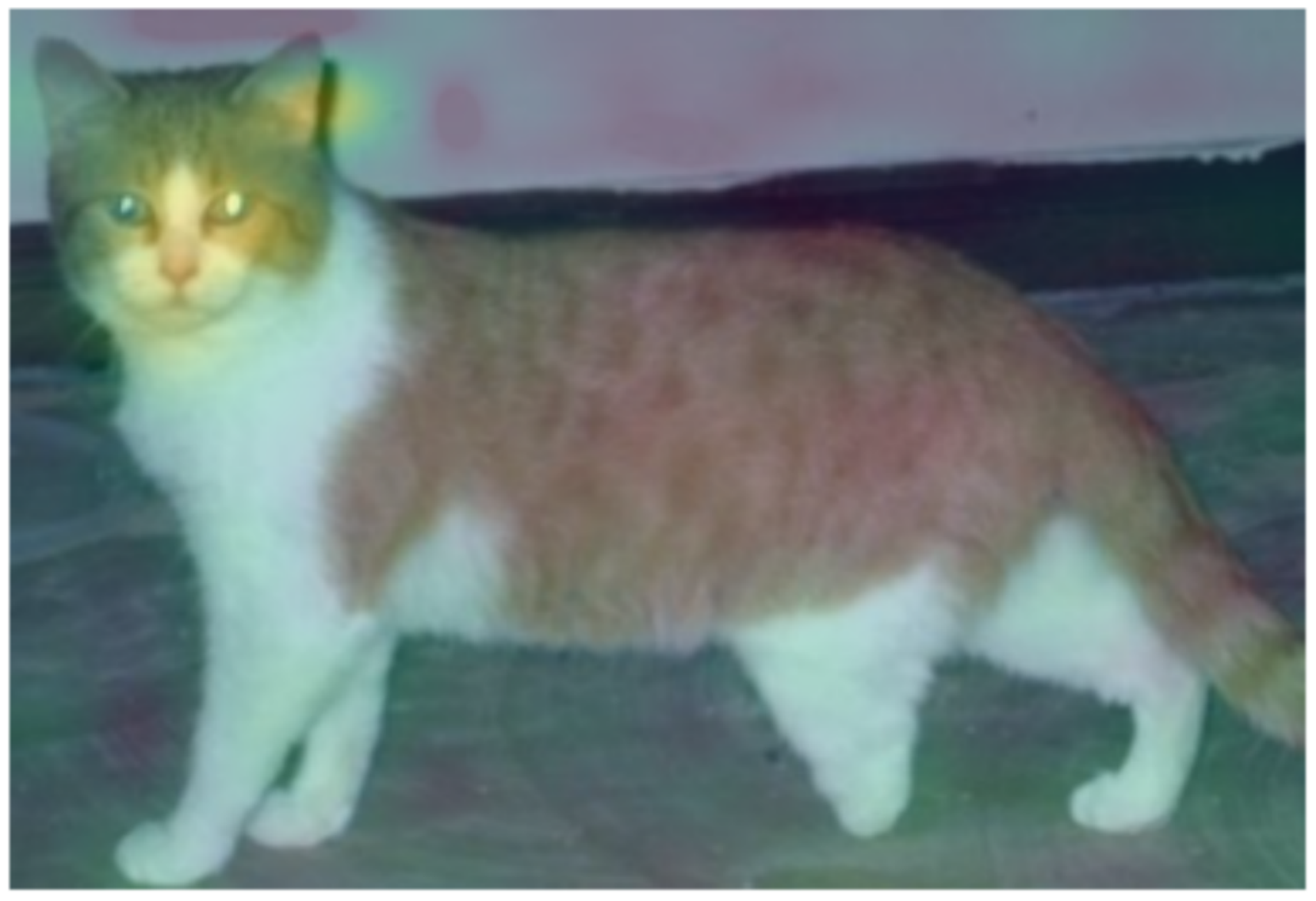} &
\includegraphics[width=0.16\textwidth]{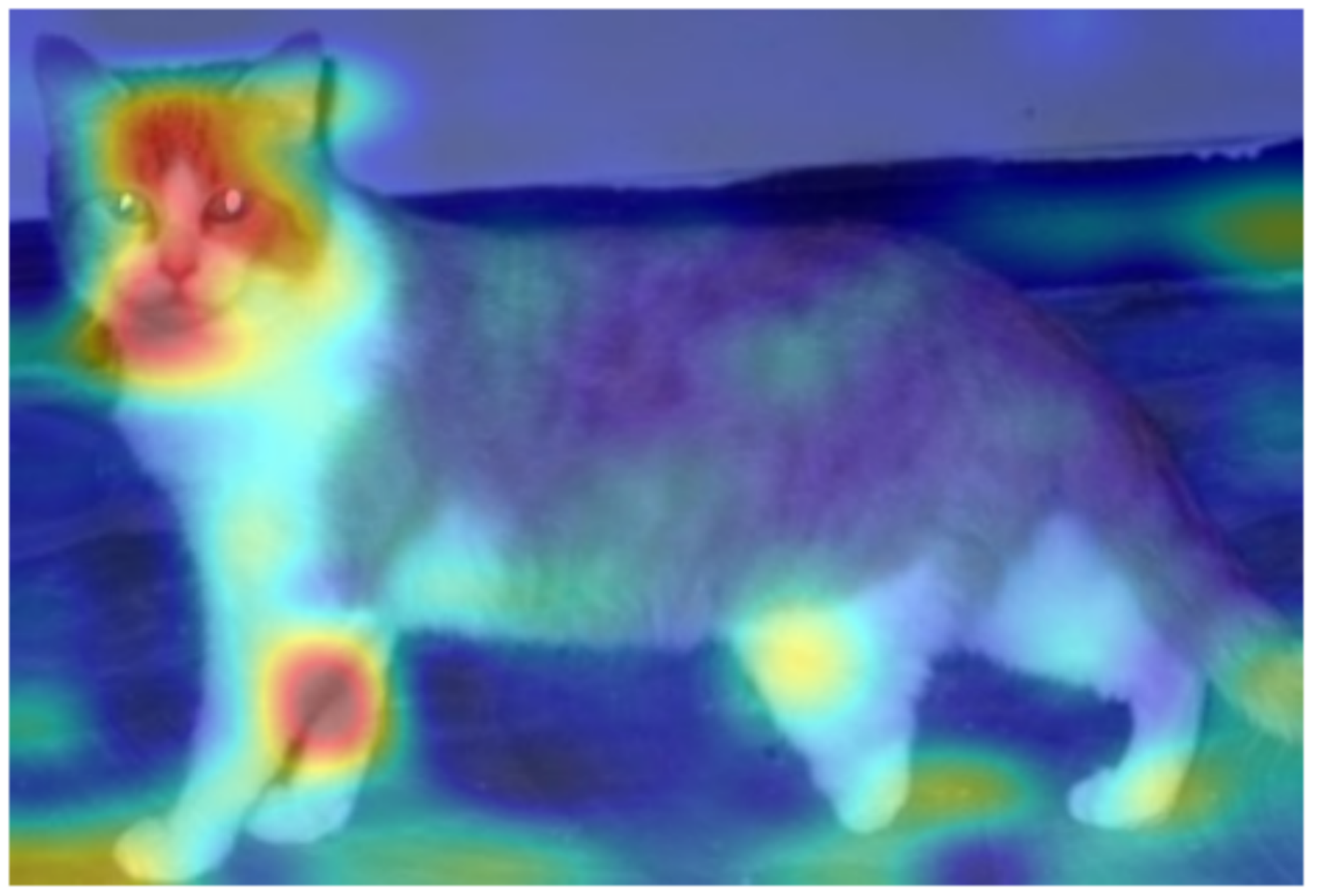} &
\includegraphics[width=0.16\textwidth]{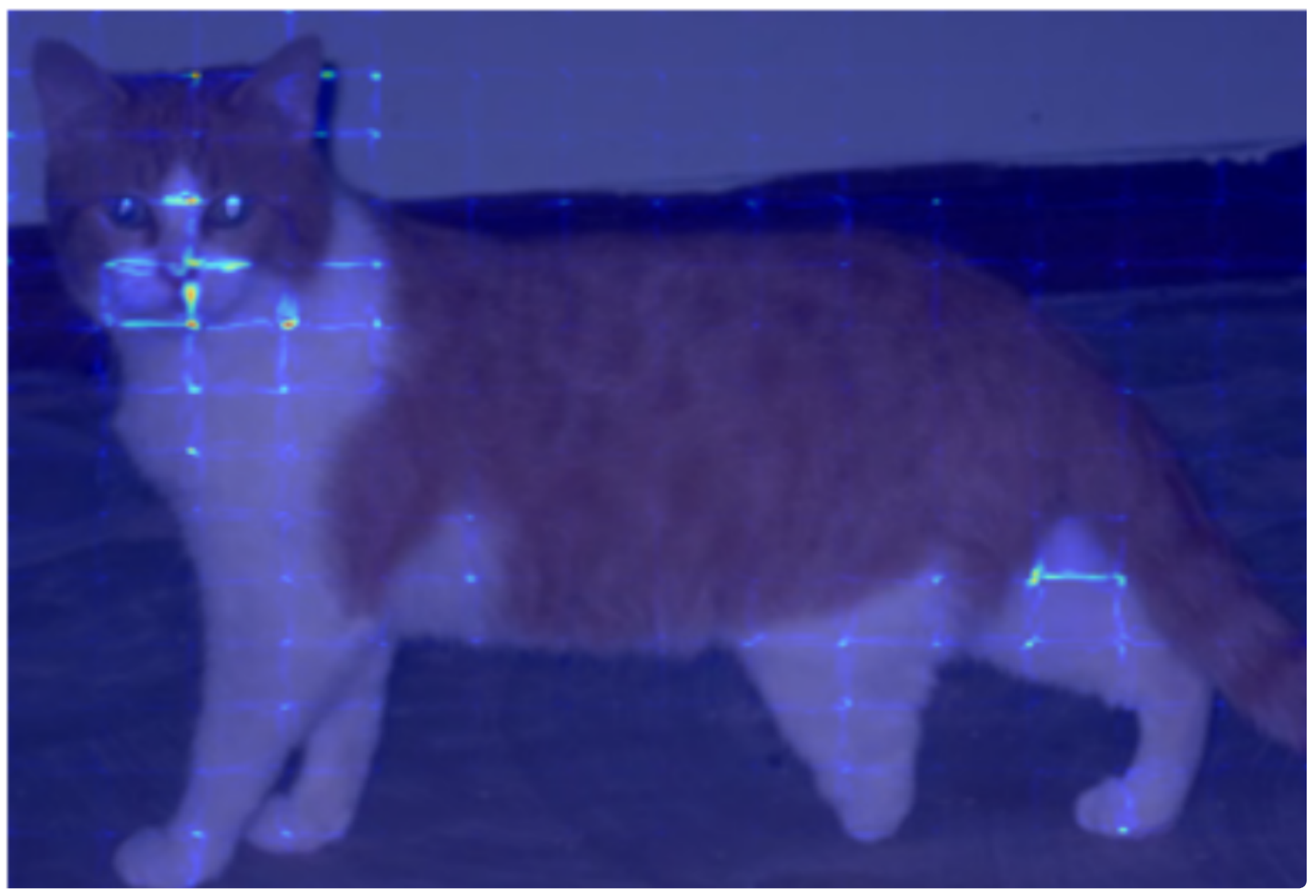} &
\includegraphics[width=0.16\textwidth]{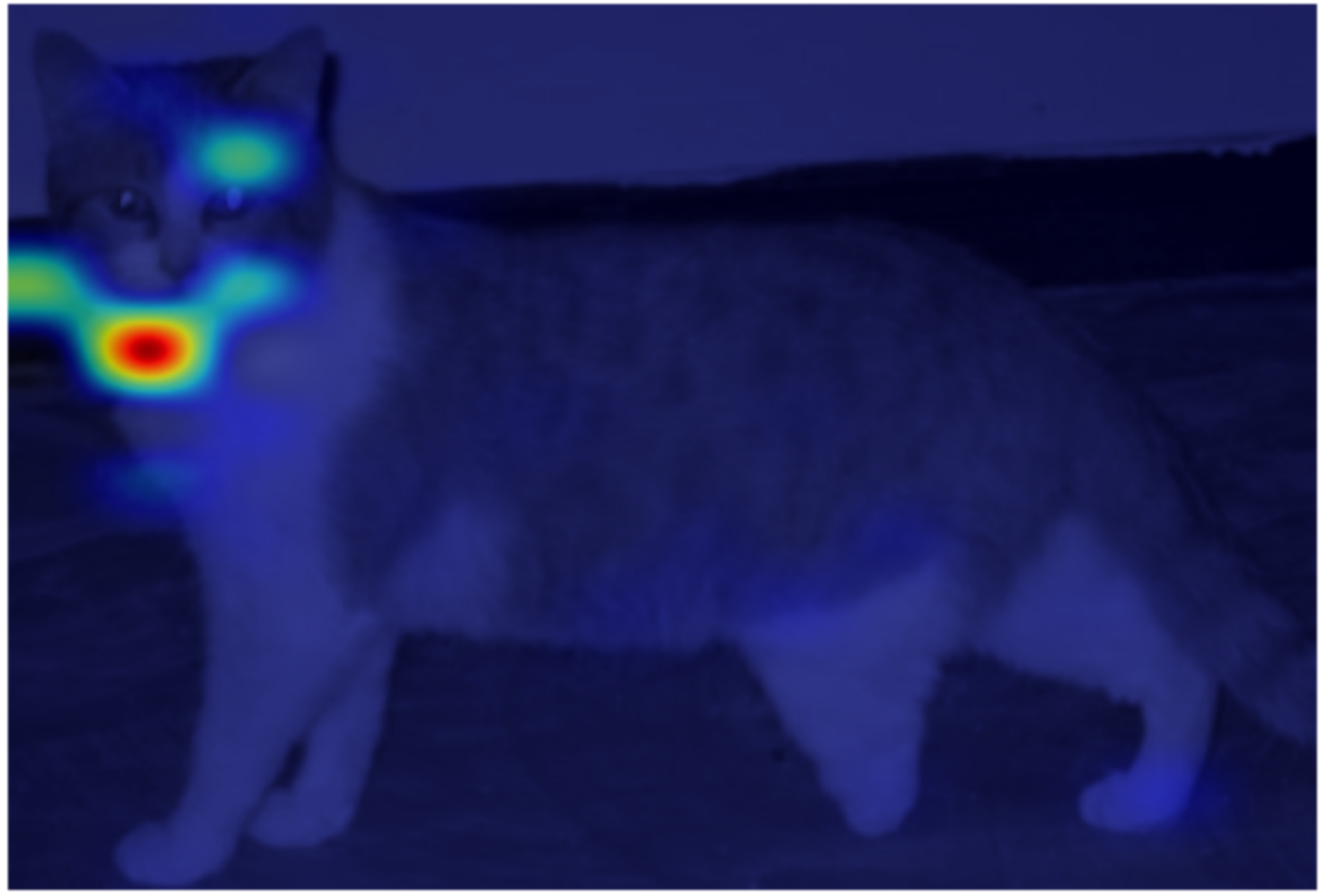} \\[2pt]

\end{tabular}

\caption{Qualitative comparison of explanations generated by various XAI methods using a Masked Autoencoder (MAE) fine-tuned on the CIFAR-10 dataset for image classification. Warmer colors denote regions with higher positive relevance to the predicted class, while cooler colors indicate lower relevance. Sample images are taken from the ASIRRA cat vs. dog dataset \cite{elson2007asirra}.}

\label{fig:baseline_comparison2}
\end{figure*}

\begin{figure*}[ht]
\centering
\includegraphics[width=\textwidth,height=0.6\textheight,keepaspectratio]{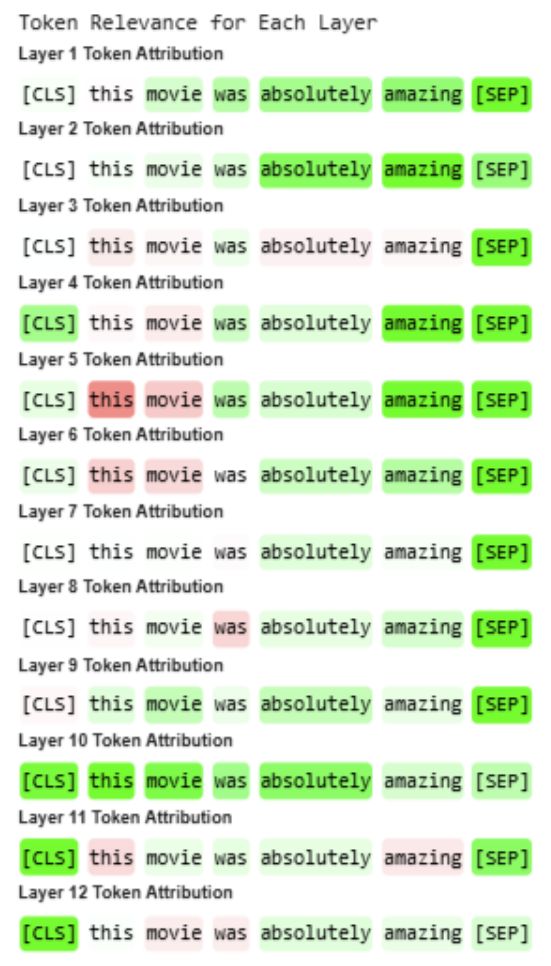}
\caption{Layer-wise token relevance visualization generated by our proposed CA-IG method using a BERT-base model fine-tuned on IMDB sentiment analysis. Brighter green indicates stronger positive evidence,
red indicates negative relevance, and white denotes neutral tokens. The progression shows early layers distributing relevance across tokens, middle layers emphasizing sentiment-bearing words (\textit{absolutely}, \textit{amazing}), and deeper layers consolidating decision-relevant cues.}
\label{Fig:Layer-wise explanation}
\end{figure*}

\twocolumn

\section{Computational Complexity Analysis}

We analyze the computational cost of the proposed CA-LIG framework in comparison with standard Integrated Gradients.
Integrated Gradients approximates the path integral between a baseline input and the actual input using $m$ interpolation steps. At each step, a gradient of the model output with respect to the input is computed. The time complexity of IG is therefore:
\begin{equation}
O(m \cdot C_{\text{grad}}),
\end{equation}
where $m$ denotes the number of interpolation steps and $C_{\text{grad}}$ represents the cost of a single gradient computation, which is approximately equivalent to one forward and one backward pass through the model. Consequently, IG explanations are roughly $m$ times more expensive than a single forward-pass inference. In practice, $m$ is typically chosen between 20 and 300, with 50 being a common default that balances numerical accuracy and computational efficiency.


The CA-IG (last-layer) variant extends IG by incorporating class-specific attention gradients as a contextual interaction signal at a single Transformer layer. While this introduces additional gradient computations for attention weights, the dominant cost remains the IG computation itself. As a result, the overall complexity remains:
\begin{equation}
O(m \cdot C_{\text{grad}}),
\end{equation}
with a small constant-factor overhead due to attention-gradient extraction.

The layer-wise approach computes IG attributions across all $L$ encoder layers of the Transformer. Since IG is applied independently at each layer, the time complexity scales linearly with the number of layers:
\begin{equation}
O(L \cdot m \cdot C_{\text{grad}}).
\end{equation}
This formulation enables hierarchical attribution analysis at the cost of increased computation.

The full \textbf{CA-LIG} framework further incorporates relevance rollout across layers by combining layer-wise IG relevance with attention-gradient-based contextual propagation. Although attention rollout itself is computationally inexpensive relative to gradient computation, CA-LIG requires computing attention gradients at each layer in addition to IG. As a result, the overall complexity remains dominated by IG and can be expressed as:
\begin{equation}
O(L \cdot m \cdot C_{\text{grad}}).
\end{equation}
The additional cost of relevance aggregation and rollout introduces only minor constant-factor overhead.

Due to its linear dependence on both the number of layers $L$ and interpolation steps $m$, CA-LIG is computationally more expensive than standard IG or other final-layer explainer methods. In practice, CA-LIG incurs approximately $L \times m$ forward--backward passes per input, making it most suitable for \textit{offline analysis}, such as model auditing, explanation benchmarking, and error analysis. 

While CA-LIG introduces additional computational overhead, this cost is justified by its ability to provide faithful, context-aware, and hierarchical explanations for Transformer-based models, thereby overcoming the limitations of existing XAI methods.

Despite a reasonable increase in computational cost, CA-LIG provides substantial advantages over conventional attribution methods. Context-aware explanations extracted from the final encoder layer are computationally efficient, as they rely on a single backward pass over already computed contextual representations, making them comparable in cost to standard Integrated Gradients. Note that IG is applied only at the final layer. Beyond this, CA-LIG enables layer-wise context-aware explanations, allowing relevance to be analyzed at different depths of the Transformer. This layered perspective reveals how token importance evolves from early lexical and syntactic representations to higher-level semantic and task-specific reasoning. As a result, CA-LIG offers richer, more faithful, and hierarchically structured explanations that better reflect the internal reasoning processes of Transformer-based models, justifying the additional computational overhead.

\clearpage

\bibliographystyle{elsarticle-num}
\bibliography{References.bib}

@article{janizek2021explaining,
  title={Explaining explanations: Axiomatic feature interactions for deep networks},
  author={Janizek, Joseph D and Sturmfels, Pascal and Lee, Su-In},
  journal={Journal of Machine Learning Research},
  volume={22},
  number={104},
  pages={1--54},
  year={2021}
}

@article{mersha2024semantic,
  title={Semantic-driven topic modeling using transformer-based embeddings and clustering algorithms},
  author={Mersha, Melkamu Abay and Kalita, Jugal and others},
  journal={Procedia Computer Science},
  volume={244},
  pages={121--132},
  year={2024},
  publisher={Elsevier}
}

@article{mersha2024ethio,
  title={Ethio-fake: Cutting-edge approaches to combat fake news in under-resourced languages using explainable ai},
  author={Mersha, Melkamu Abay and Bade, Girma Yohannis and Kalita, Jugal and Kolesnikova, Olga and Gelbukh, Alexander and others},
  journal={Procedia Computer Science},
  volume={244},
  pages={133--142},
  year={2024},
  publisher={Elsevier}
}

@article{khapre2025toxicity,
  title={Toxicity in online platforms and AI systems: A survey of needs, challenges, mitigations, and future directions},
  author={Khapre, Smita and Mersha, Melkamu Abay and Shakil, Hassan and Baruah, Jonali and Kalita, Jugal},
  journal={Expert Systems with Applications},
  pages={129832},
  year={2025},
  publisher={Elsevier}
}

@inproceedings{tonja2023first,
  title={First attempt at building parallel corpora for machine translation of northeast india’s very low-resource languages},
  author={Tonja, Atnafu Lambebo and Mersha, Melkamu and Kalita, Ananya and Kolesnikova, Olga and Kalita, Jugal},
  booktitle={Proceedings of the 20th International Conference on Natural Language Processing (ICON)},
  pages={534--539},
  year={2023}
}

@inproceedings{kamen2025introducing,
  title={Introducing Semantic Feature Dependencies in NLP XAI Systems with SUPLIME},
  author={Kamen, Daniel and Mersha, Melkamu Abay and Kalita, Jugal},
  booktitle={Recent Advances in Natural Language Processing},
  pages={47},
  year={2025}
}

@article{mersha2025semantic,
  title={Semantic-driven topic modeling for analyzing creativity in virtual brainstorming},
  author={Mersha, Melkamu Abay and Kalita, Jugal},
  journal={arXiv preprint arXiv:2509.16835},
  year={2025}
}

@article{alshami2025smart,
  title={Smart-vision: survey of modern action recognition techniques in vision},
  author={AlShami, Ali K and Rabinowitz, Ryan and Lam, Khang and Shleibik, Yousra and Mersha, Melkamu and Boult, Terrance and Kalita, Jugal},
  journal={Multimedia tools and applications},
  volume={84},
  number={27},
  pages={32705--32776},
  year={2025},
  publisher={Springer}
}

@article{mersha2025unified,
  title={A unified framework with novel metrics for evaluating the effectiveness of xai techniques in llms},
  author={Mersha, Melkamu Abay and Yigezu, Mesay Gemeda and Shakil, Hassan and AlShami, Ali K and Byun, Sanghyun and Kalita, Jugal},
  journal={arXiv preprint arXiv:2503.05050},
  year={2025}
}

@article{mersha2024explainability,
  title={Explainability in neural networks for natural language processing tasks},
  author={Mersha, Melkamu and Bitewa, Mingiziem and Abay, Tsion and Kalita, Jugal},
  journal={arXiv preprint arXiv:2412.18036},
  year={2024}
}

@article{rogers2020primer,
  title={A primer in BERTology: What we know about how BERT works},
  author={Rogers, Anna and Kovaleva, Olga and Rumshisky, Anna},
  journal={Transactions of the association for computational linguistics},
  volume={8},
  pages={842--866},
  year={2020}
}

@article{ferrando2024primer,
  title={A primer on the inner workings of transformer-based language models},
  author={Ferrando, Javier and Sarti, Gabriele and Bisazza, Arianna and Costa-Juss{\`a}, Marta R},
  journal={arXiv preprint arXiv:2405.00208},
  year={2024}
}

@inproceedings{zhu2018hierarchical,
  title={Hierarchical attention flow for multiple-choice reading comprehension},
  author={Zhu, Haichao and Wei, Furu and Qin, Bing and Liu, Ting},
  booktitle={Proceedings of the AAAI Conference on Artificial Intelligence},
  volume={32},
  number={1},
  year={2018}
}

@article{lan2025attention,
  title={Attention Consistency for LLMs Explanation},
  author={Lan, Tian and Xu, Jinyuan and He, Xue and Hwang, Jenq-Neng and Li, Lei},
  journal={arXiv preprint arXiv:2509.17178},
  year={2025}
}

@article{achtibat2024attnlrp,
  title={Attnlrp: attention-aware layer-wise relevance propagation for transformers},
  author={Achtibat, Reduan and Hatefi, Sayed Mohammad Vakilzadeh and Dreyer, Maximilian and Jain, Aakriti and Wiegand, Thomas and Lapuschkin, Sebastian and Samek, Wojciech},
  journal={arXiv preprint arXiv:2402.05602},
  year={2024}
}

@article{sood2020interpreting,
  title={Interpreting attention models with human visual attention in machine reading comprehension},
  author={Sood, Ekta and Tannert, Simon and Frassinelli, Diego and Bulling, Andreas and Vu, Ngoc Thang},
  journal={arXiv preprint arXiv:2010.06396},
  year={2020}
}

@inproceedings{rogers2020,
  title        = {A Primer in {BERT}ology: What we know about how {BERT} works},
  author       = {Rogers, Anna and Kovaleva, Olga and Rumshisky, Anna},
  booktitle    = {Proceedings of the 58th Annual Meeting of the Association for Computational Linguistics},
  year         = {2020},
  pages        = {1--17},
 }

@article{conneau2019unsupervised,
  title={Unsupervised cross-lingual representation learning at scale},
  author={Conneau, Alexis and Khandelwal, Kartikay and Goyal, Naman and Chaudhary, Vishrav and Wenzek, Guillaume and Guzm{\'a}n, Francisco and Grave, Edouard and Ott, Myle and Zettlemoyer, Luke and Stoyanov, Veselin},
  journal={arXiv preprint arXiv:1911.02116},
  year={2019}
}

@article{krizhevsky2009learning,
  title={Learning multiple layers of features from tiny images},
  author={Krizhevsky, Alex and Hinton, Geoffrey and others},
  year={2009},
  publisher={Toronto, ON, Canada}
}

@article{abnar2020quantifying,
  title={Quantifying attention flow in transformers},
  author={Abnar, Samira and Zuidema, Willem},
  journal={arXiv preprint arXiv:2005.00928},
  year={2020}
}

@article{han2025contrast,
  title={Contrast-CAT: Contrasting Activations for Enhanced Interpretability in Transformer-based Text Classifiers},
  author={Han, Sungmin and Lee, Jeonghyun and Lee, Sangkyun},
  journal={arXiv preprint arXiv:2507.21186},
  year={2025}
}

@article{shi2024ircan,
  title={Ircan: Mitigating knowledge conflicts in llm generation via identifying and reweighting context-aware neurons},
  author={Shi, Dan and Jin, Renren and Shen, Tianhao and Dong, Weilong and Wu, Xinwei and Xiong, Deyi},
  journal={Advances in Neural Information Processing Systems},
  volume={37},
  pages={4997--5024},
  year={2024}
}

@inproceedings{azarkhalili2025generalized,
  title={Generalized Attention Flow: Feature Attribution for Transformer Models via Maximum Flow},
  author={Azarkhalili, Behrooz and Libbrecht, Maxwell W},
  booktitle={Proceedings of the 63rd Annual Meeting of the Association for Computational Linguistics (Volume 1: Long Papers)},
  pages={19954--19974},
  year={2025}
}

@article{mersha2025evaluating,
  title={Evaluating the effectiveness of XAI techniques for encoder-based language models},
  author={Mersha, Melkamu Abay and Yigezu, Mesay Gemeda and Kalita, Jugal},
  journal={Knowledge-Based Systems},
  volume={310},
  pages={113042},
  year={2025},
  publisher={Elsevier}
}

@inproceedings{peters2018,
  title        = {Deep Contextualized Word Representations},
  author       = {Peters, Matthew E. and Neumann, Mark and Iyyer, Mohit and Gardner, Matt and Clark, Christopher and Lee, Kenton and Zettlemoyer, Luke},
  booktitle    = {Proceedings of the 2018 Conference of the North American Chapter of the Association for Computational Linguistics: Human Language Technologies},
  year         = {2018},
  pages        = {2227--2237},
  }

@inproceedings{selvaraju2017grad,
  title={Grad-cam: Visual explanations from deep networks via gradient-based localization},
  author={Selvaraju, Ramprasaath R and Cogswell, Michael and Das, Abhishek and Vedantam, Ramakrishna and Parikh, Devi and Batra, Dhruv},
  booktitle={Proceedings of the IEEE international conference on computer vision},
  pages={618--626},
  year={2017}
}

@article{elson2007asirra,
  title={Asirra: a CAPTCHA that exploits interest-aligned manual image categorization.},
  author={Elson, Jeremy and Douceur, John R and Howell, Jon and Saul, Jared},
  journal={CCS},
  volume={7},
  number={366-374},
  pages={15},
  year={2007}
}

@inproceedings{hewitt2019structural,
  title     = {A Structural Probe for Finding Syntax in Word Representations},
  author    = {Hewitt, John and Manning, Christopher D.},
  booktitle = {Proceedings of the 2019 Conference of the North American Chapter of the Association for Computational Linguistics: Human Language Technologies, Volume 1 (Long and Short Papers)},
  year      = {2019},
  pages     = {4129--4138},
  publisher = {Association for Computational Linguistics},
  }

@inproceedings{goldberg2019assessing,
  title     = {Assessing {BERT}'s Syntactic Abilities},
  author    = {Goldberg, Yoav},
  booktitle = {Proceedings of the 57th Annual Meeting of the Association for Computational Linguistics},
  year      = {2019},
  pages     = {3623--3632},
  publisher = {Association for Computational Linguistics},
  }

@inproceedings{liu2019linguistic,
  title     = {Linguistic Knowledge and Transferability of Contextual Representations},
  author    = {Liu, Nelson F. and Gardner, Matt and Belinkov, Yonatan and Peters, Matthew and Smith, Noah A.},
  booktitle = {Proceedings of the 2019 Conference of the North American Chapter of the Association for Computational Linguistics: Human Language Technologies, Volume 1 (Long and Short Papers)},
  year      = {2019},
  pages     = {1073--1094},
  publisher = {Association for Computational Linguistics},
  }

@inproceedings{clark2019bert,
  title     = {What Does {BERT} Look at? An Analysis of {BERT}'s Attention},
  author    = {Clark, Kevin and Khandelwal, Urvashi and Levy, Omer and Manning, Christopher D.},
  booktitle = {Proceedings of the 2019 ACL Workshop BlackboxNLP: Analyzing and Interpreting Neural Networks for NLP},
  year      = {2019},
  pages     = {276--286},
  publisher = {Association for Computational Linguistics},
  }

@inproceedings{sun2019fine,
  title     = {Fine-Tune {BERT} for Extractive Summarization},
  author    = {Sun, Chi and Qiu, Xipeng and Xu, Yige and Huang, Xuanjing},
  booktitle = {Proceedings of the 2019 Conference on Empirical Methods in Natural Language Processing and the 9th International Joint Conference on Natural Language Processing (EMNLP-IJCNLP)},
  year      = {2019},
  pages     = {3289--3299},
  publisher = {Association for Computational Linguistics},
  }

@inproceedings{ethayarajh2019contextual,
  title     = {How Contextual are Contextualized Word Representations? Comparing the Geometry of {BERT}, {ELMo}, and {GPT}-2 Embeddings},
  author    = {Ethayarajh, Kawin},
  booktitle = {Proceedings of the 2019 Conference on Empirical Methods in Natural Language Processing and the 9th International Joint Conference on Natural Language Processing (EMNLP-IJCNLP)},
  year      = {2019},
  pages     = {55--65},
  publisher = {Association for Computational Linguistics},
}

@inproceedings{kovaleva2019revealing,
  title     = {Revealing the Dark Secrets of {BERT}},
  author    = {Kovaleva, Olga and Romanov, Alexey and Rogers, Anna and Rumshisky, Anna},
  booktitle = {Proceedings of the 2019 Conference on Empirical Methods in Natural Language Processing and the 9th International Joint Conference on Natural Language Processing (EMNLP-IJCNLP)},
  year      = {2019},
  pages     = {4365--4374},
  publisher = {Association for Computational Linguistics},
  }

@article{arras2017explaining,
  title={Explaining recurrent neural network predictions in sentiment analysis},
  author={Arras, Leila and Montavon, Gr{\'e}goire and M{\"u}ller, Klaus-Robert and Samek, Wojciech},
  journal={arXiv preprint arXiv:1706.07206},
  year={2017}
}

@inproceedings{maas2011learning,
  title={Learning word vectors for sentiment analysis},
  author={Maas, Andrew and Daly, Raymond E and Pham, Peter T and Huang, Dan and Ng, Andrew Y and Potts, Christopher},
  booktitle={Proceedings of the 49th annual meeting of the association for computational linguistics: Human language technologies},
  pages={142--150},
  year={2011}
}

@inproceedings{tenney2019bert,
  title     = {{BERT} Rediscovers the Classical {NLP} Pipeline},
  author    = {Tenney, Ian and Das, Dipanjan and Pavlick, Ellie},
  booktitle = {Proceedings of the 57th Annual Meeting of the Association for Computational Linguistics},
  year      = {2019},
  pages     = {4593--4601},
  publisher = {Association for Computational Linguistics},
  }

@inproceedings{hollenstein2021relative,
  title={Relative importance in sentence processing},
  author={Hollenstein, Nora and Beinborn, Lisa},
  booktitle={Proceedings of the 59th Annual Meeting of the Association for Computational Linguistics and the 11th International Joint Conference on Natural Language Processing (ACL-IJCNLP)},
  pages={141--150},
  year={2021}
}

@inproceedings{shrikumar2017learning,
  title={Learning important features through propagating activation differences},
  author={Shrikumar, Avanti and Greenside, Peyton and Kundaje, Anshul},
  booktitle={International conference on machine learning},
  pages={3145--3153},
  year={2017},
  organization={PMlR}
}

@inproceedings{ayele2023exploring,
  title={Exploring Amharic hate speech data collection and classification approaches},
  author={Ayele, Abinew Ali and Yimam, Seid Muhie and Belay, Tadesse Destaw and Asfaw, Tesfa and Biemann, Chris},
  booktitle={Proceedings of the 14th international conference on recent advances in natural language processing},
  pages={49--59},
  year={2023}
}

@inproceedings{hou2023decoding,
  title={Decoding layer saliency in language transformers},
  author={Hou, Elizabeth Mary and Castanon, Gregory David},
  booktitle={International Conference on Machine Learning},
  pages={13285--13308},
  year={2023},
  organization={PMLR}
}

@article{fantozzi2024explainability,
  title={Explainability in Deep Learning: Challenges for Transformers},
  author={Fantozzi, Matteo and others},
  journal={Frontiers in Artificial Intelligence},
  year={2024}
}

@inproceedings{sundararajan2017axiomatic,
  title={Axiomatic Attribution for Deep Networks},
  author={Sundararajan, Mukund and Taly, Ankur and Yan, Qiqi},
  booktitle={ICML},
  year={2017}
}

@article{smilkov2017smoothgrad,
  title={SmoothGrad: removing noise by adding noise},
  author={Smilkov, Daniel and others},
  journal={arXiv preprint arXiv:1706.03825},
  year={2017}
}

@inproceedings{zaidan2007using,
  title={Using “annotator rationales” to improve machine learning for text categorization},
  author={Zaidan, Omar and Eisner, Jason and Piatko, Christine},
  booktitle={Human language technologies 2007: The conference of the North American chapter of the association for computational linguistics; proceedings of the main conference},
  pages={260--267},
  year={2007}
}

@article{deyoung2019eraser,
  title={ERASER: A benchmark to evaluate rationalized NLP models},
  author={DeYoung, Jay and Jain, Sarthak and Rajani, Nazneen Fatema and Lehman, Eric and Xiong, Caiming and Socher, Richard and Wallace, Byron C},
  journal={arXiv preprint arXiv:1911.03429},
  year={2019}
}

@inproceedings{he2022masked,
  title={Masked autoencoders are scalable vision learners},
  author={He, Kaiming and Chen, Xinlei and Xie, Saining and Li, Yanghao and Doll{\'a}r, Piotr and Girshick, Ross},
  booktitle={Proceedings of the IEEE/CVF conference on computer vision and pattern recognition},
  pages={16000--16009},
  year={2022}
}

@inproceedings{devlin2019bert,
  title={Bert: Pre-training of deep bidirectional transformers for language understanding},
  author={Devlin, Jacob and Chang, Ming-Wei and Lee, Kenton and Toutanova, Kristina},
  booktitle={Proceedings of the 2019 conference of the North American chapter of the association for computational linguistics: human language technologies, volume 1 (long and short papers)},
  pages={4171--4186},
  year={2019}
}

@article{chen2024interpretable,
  title={An interpretable and transferrable vision transformer model for rapid materials spectra classification},
  author={Chen, Zhenru and Xie, Yunchao and Wu, Yuchao and Lin, Yuyi and Tomiya, Shigetaka and Lin, Jian},
  journal={Digital Discovery},
  volume={3},
  number={2},
  pages={369--380},
  year={2024},
  publisher={Royal Society of Chemistry}
}

@article{jain2019attention,
  title={Attention is not explanation},
  author={Jain, Sarthak and Wallace, Byron C},
  journal={arXiv preprint arXiv:1902.10186},
  year={2019}
}

@article{wiegreffe2019attention,
  title={Attention is not not explanation},
  author={Wiegreffe, Sarah and Pinter, Yuval},
  journal={arXiv preprint arXiv:1908.04626},
  year={2019}
}

@article{serrano2019attention,
  title={Is attention interpretable?},
  author={Serrano, Sofia and Smith, Noah A},
  journal={arXiv preprint arXiv:1906.03731},
  year={2019}
}

@misc{jain2023inseq,
  title={Inseq: A Toolkit for Sequence-Level Interpretability of NLP Models},
  author={Jain, Sarthak and others},
  year={2023},
  howpublished={\url{https://github.com/penwang/inseq}}
}

@inproceedings{chefer2021transformer,
  title={Transformer Interpretability Beyond Attention Visualization},
  author={Chefer, Hila and Gur, Shir and Wolf, Lior},
  booktitle={CVPR},
  year={2021}
}

@inproceedings{ali2022xai,
  title={XAI Methods for Transformers via Conservative Propagation},
  author={Ali, Ahmed and Kumar, Aditya},
  booktitle={ICLR},
  year={2022}
}

@incollection{lang1995newsweeder,
  title={Newsweeder: Learning to filter netnews},
  author={Lang, Ken},
  booktitle={Machine learning proceedings 1995},
  pages={331--339},
  year={1995},
  publisher={Elsevier}
}

@inproceedings{dossou2022afrolm,
  title={AfroLM: A self-active learning-based multilingual pretrained language model for 23 African languages},
  author={Dossou, Bonaventure FP and Tonja, Atnafu Lambebo and Yousuf, Oreen and Osei, Salomey and Oppong, Abigail and Shode, Iyanuoluwa and Awoyomi, Oluwabusayo Olufunke and Emezue, Chris},
  booktitle={Proceedings of The Third Workshop on Simple and Efficient Natural Language Processing (SustaiNLP)},
  pages={52--64},
  year={2022}
}

@article{mersha2025explainable,
  title={Explainable AI: XAI-Guided Context-Aware Data Augmentation},
  author={Mersha, Melkamu Abay and Yigezu, Mesay Gemeda and Tonja, Atnafu Lambebo and Shakil, Hassan and Iskandar, Samer and Kolesnikova, Olga and Kalita, Jugal},
  journal={Expert Systems with Applications},
  pages={128364},
  year={2025},
  publisher={Elsevier}
}

@inproceedings{ribeiro2016should,
  title={" Why should i trust you?" Explaining the predictions of any classifier},
  author={Ribeiro, Marco Tulio and Singh, Sameer and Guestrin, Carlos},
  booktitle={Proceedings of the 22nd ACM SIGKDD international conference on knowledge discovery and data mining},
  pages={1135--1144},
  year={2016}
}

@article{lundberg2017unified,
  title={A unified approach to interpreting model predictions},
  author={Lundberg, Scott},
  journal={arXiv preprint arXiv:1705.07874},
  year={2017}
}

@inproceedings{zeiler2014visualizing,
  title={Visualizing and Understanding Convolutional Networks},
  author={Zeiler, MD},
  booktitle={European conference on computer vision/arXiv},
  volume={1311},
  year={2014}
}

@inproceedings{zhou2016learning,
  title={Learning deep features for discriminative localization},
  author={Zhou, Bolei and Khosla, Aditya and Lapedriza, Agata and Oliva, Aude and Torralba, Antonio},
  booktitle={Proceedings of the IEEE conference on computer vision and pattern recognition},
  pages={2921--2929},
  year={2016}
}

@article{bach2015pixel,
  title={On pixel-wise explanations for non-linear classifier decisions by layer-wise relevance propagation},
  author={Bach, Sebastian and Binder, Alexander and Montavon, Gr{\'e}goire and Klauschen, Frederick and M{\"u}ller, Klaus-Robert and Samek, Wojciech},
  journal={PloS one},
  volume={10},
  number={7},
  pages={e0130140},
  year={2015},
  publisher={Public Library of Science San Francisco, CA USA}
}

@inproceedings{kim2018interpretability,
  title={Interpretability beyond feature attribution: Quantitative testing with concept activation vectors (tcav)},
  author={Kim, Been and Wattenberg, Martin and Gilmer, Justin and Cai, Carrie and Wexler, James and Viegas, Fernanda and others},
  booktitle={International conference on machine learning},
  pages={2668--2677},
  year={2018},
  organization={PMLR}
}

@article{srinivas2019full,
  title={Full-gradient representation for neural network visualization},
  author={Srinivas, Suraj and Fleuret, Fran{\c{c}}ois},
  journal={Advances in neural information processing systems},
  volume={32},
  year={2019}
}

@inproceedings{kapishnikov2021guided,
  title={Guided integrated gradients: An adaptive path method for removing noise},
  author={Kapishnikov, Andrei and Venugopalan, Subhashini and Avci, Besim and Wedin, Ben and Terry, Michael and Bolukbasi, Tolga},
  booktitle={Proceedings of the IEEE/CVF conference on computer vision and pattern recognition},
  pages={5050--5058},
  year={2021}
}

@article{vig2019multiscale,
  title={A multiscale visualization of attention in the transformer model},
  author={Vig, Jesse},
  journal={arXiv preprint arXiv:1906.05714},
  year={2019}
}

@article{selvaraju2020grad,
  title={Grad-CAM: visual explanations from deep networks via gradient-based localization},
  author={Selvaraju, Ramprasaath R and Cogswell, Michael and Das, Abhishek and Vedantam, Ramakrishna and Parikh, Devi and Batra, Dhruv},
  journal={International journal of computer vision},
  volume={128},
  pages={336--359},
  year={2020},
  publisher={Springer}
}

@article{qiang2022attcat,
  title={Attcat: Explaining transformers via attentive class activation tokens},
  author={Qiang, Yao and Pan, Deng and Li, Chengyin and Li, Xin and Jang, Rhongho and Zhu, Dongxiao},
  journal={Advances in neural information processing systems},
  volume={35},
  pages={5052--5064},
  year={2022}
}

@inproceedings{yuan2021explaining,
  title={Explaining information flow inside vision transformers using markov chain},
  author={Yuan, Tingyi and Li, Xuhong and Xiong, Haoyi and Cao, Hui and Dou, Dejing},
  booktitle={eXplainable AI approaches for debugging and diagnosis.},
  year={2021}
}

@article{raffel2020exploring,
  title={Exploring the limits of transfer learning with a unified text-to-text transformer},
  author={Raffel, Colin and Shazeer, Noam and Roberts, Adam and Lee, Katherine and Narang, Sharan and Matena, Michael and Zhou, Yanqi and Li, Wei and Liu, Peter J},
  journal={Journal of machine learning research},
  volume={21},
  number={140},
  pages={1--67},
  year={2020}
}

@article{vaswani2017attention,
  title={Attention is all you need},
  author={Vaswani, A},
  journal={Advances in Neural Information Processing Systems},
  year={2017}
}

@inproceedings{liu2021exploring,
  title={On exploring attention-based explanation for transformer models in text classification},
  author={Liu, Shengzhong and Le, Franck and Chakraborty, Supriyo and Abdelzaher, Tarek},
  booktitle={2021 IEEE International Conference on Big Data (Big Data)},
  pages={1193--1203},
  year={2021},
  organization={IEEE}
}

@article{yeh2023attentionviz,
  title={Attentionviz: A global view of transformer attention},
  author={Yeh, Catherine and Chen, Yida and Wu, Aoyu and Chen, Cynthia and Vi{\'e}gas, Fernanda and Wattenberg, Martin},
  journal={IEEE Transactions on Visualization and Computer Graphics},
  year={2023},
  publisher={IEEE}
}

@article{devlin2018bert,
  title={Bert: Pre-training of deep bidirectional transformers for language understanding},
  author={Devlin, Jacob and Chang, Ming-Wei and Lee, Kenton and Toutanova, Kristina},
  journal={arXiv preprint arXiv:1810.04805},
  year={2018}
}

@article{radford2018improving,
  title={Improving language understanding by generative pre-training},
  author={Radford, Alec and Narasimhan, Karthik and Salimans, Tim and Sutskever, Ilya and others},
  year={2018},
  publisher={OpenAI}
}

@inproceedings{aoyama2022probe,
  title     = {Probe-Less Probing of {BERT}'s Layer-Wise Linguistic Knowledge with Masked Word Prediction},
  author    = {Aoyama, Tatsuya and Schneider, Nathan},
  booktitle = {Proceedings of the 2022 Conference of the North American Chapter of the Association for Computational Linguistics: Student Research Workshop},
  year      = {2022},
  pages     = {195--201},
  publisher = {Association for Computational Linguistics},
  }

@article{liu2024cunliang,
  title={Cunliang Kong, Ying Liu, and Maosong Sun. 2024. Fantastic semantics and where to find them: Investigating which layers of generative LLMs reflect lexical semantics},
  author={Liu, Zhu},
  journal={Findings of the Association for Computational Linguistics: ACL},
  pages={14551--14558},
  year={2024}
}

@inproceedings{ferrando2022measure,
  title     = {Measuring the Mixing of Contextual Information in the Transformer},
  author    = {Ferrando, Jorge},
  booktitle = {Proceedings of the 2022 Conference on Empirical Methods in Natural Language Processing},
  year      = {2022},
  publisher = {Association for Computational Linguistics},
 }

@article{nauta2023anecdotal,
  title={From anecdotal evidence to quantitative evaluation methods: A systematic review on evaluating explainable ai},
  author={Nauta, Meike and Trienes, Jan and Pathak, Shreyasi and Nguyen, Elisa and Peters, Michelle and Schmitt, Yasmin and Schl{\"o}tterer, J{\"o}rg and van Keulen, Maurice and Seifert, Christin},
  journal={ACM Computing Surveys},
  volume={55},
  number={13s},
  pages={1--42},
  year={2023},
  publisher={ACM New York, NY}
}

@article{mersha2024explainable,
  title={Explainable artificial intelligence: A survey of needs, techniques, applications, and future direction},
  author={Mersha, Melkamu and Lam, Khang and Wood, Joseph and AlShami, Ali and Kalita, Jugal},
  journal={Neurocomputing},
  pages={128111},
  year={2024},
  publisher={Elsevier}
}

\end{document}